\documentclass{ieeeaccess}
\usepackage{cite}
\usepackage{amsmath,amssymb,amsfonts}
\usepackage{amsthm}
\usepackage{graphicx}
\usepackage{textcomp}
\usepackage{amsmath}
\usepackage{bbm}
\usepackage[numbers]{natbib}
\usepackage{algorithm}
\usepackage[noend]{algpseudocode}
\usepackage[T1]{fontenc}
\usepackage[utf8]{inputenc}
\usepackage{multirow}
\usepackage{hyperref}
\usepackage{url}
\usepackage{amsfonts}
\usepackage{longtable} 
\usepackage{booktabs}
\usepackage{subfig}
\usepackage[misc,geometry]{ifsym} 
\usepackage[labelfont=bf,textfont=md]{caption}
\def\BibTeX{{\rm B\kern-.05em{\sc i\kern-.025em b}\kern-.08em
    T\kern-.1667em\lower.7ex\hbox{E}\kern-.125emX}}
    
\newtheorem{definition}{Definition}   
    
\renewcommand{\tabcolsep}{1mm}
    
\begin{document}

\history{Date of publication xxxx 00, 0000, date of current version xxxx 00, 0000.}
\doi{10.1109/ACCESS.2017.DOI}

\title{Approaches For Multi-View Redescription Mining}
\author{\uppercase{Matej Mihel\v{c}i\'{c}}\authorrefmark{1}, and \uppercase{Tomislav \v{S}muc}\authorrefmark{2}}
\address[1]{Department of Mathematics, Faculty of Science, Bijenička cesta 30, 10000 Zagreb, Croatia (e-mail: matmih@math.hr)}
\address[2]{Ruđer Bošković Institute, Bijenička cesta 54, 10000 Zagreb, Croatia (e-mail: smuc@irb.hr)}

\markboth
{Mihelčić \headeretal: Approaches for multi-view redescription mining}
{Mihelčić \headeretal: Approaches for multi-view redescription mining}

\corresp{Corresponding author: Matej Mihelčić (e-mail: matmih@math.hr).}

\begin{abstract}
The task of redescription mining explores ways to re-describe different subsets of entities contained in a dataset and to reveal non-trivial associations between different subsets of attributes, called views. This interesting and challenging task is encountered in different scientific fields, and is addressed by a number of approaches that obtain redescriptions and allow for the exploration and analyses of attribute associations. The main limitation of existing approaches to this task is their inability to use more than two views. Our work alleviates this drawback. We present a memory efficient, extensible multi-view redescription mining framework that can be used to relate multiple, i.e. more than two views, disjoint sets of attributes describing one set of entities. The framework can use any multi-target regression or multi-label classification algorithm, with models that can be represented as sets of rules, to generate redescriptions. Multi-view redescriptions are built using incremental view-extending heuristic from initially created two-view redescriptions. In this work, we use different types of Predictive Clustering trees algorithms (regular, extra, with random output selection) and the Random Forest thereof in order to improve the quality of final redescription sets and/or execution time needed to generate them. We provide multiple performance analyses of the proposed framework and compare it against the naive approach to multi-view redescription mining.  We demonstrate the usefulness of the proposed multi-view extension on several datasets, including a use-case on understanding of machine learning models - a topic of growing importance in machine learning and artificial intelligence in general.

\end{abstract}

\begin{keywords}
knowledge discovery, multi-view redescription mining, redescription set, predictive clustering trees, random forest, extremely randomized trees, random output selection.
\end{keywords}

\titlepgskip=-15pt

\maketitle

\section{Introduction}
\label{intro}

Redescription mining \cite{Ramakrishnan:2004} aims to find multiple  characterizations (re-descriptions) of different subsets of entities in a set of available data, i.e. to identify subsets of instances that can be re-described. These characterisations are expressed in a rule-like form which makes them easy to understand. Entities can be re-described using, e.g. information about entities obtained from different data sources or using different types of data, using data describing different aspects of the entities, or  describing entities at different points in time. The result of analysis performed by redescription mining is a set of redescriptions - tuples of logical formulas (also called queries), where queries of a redescription depict the same, or very similar subsets of entities - the intersection of which is called a redescription support set.

The ability of redescription mining to provide tuples of rules re-describing entities in a support set (using multiple sources of data, used to form views) distinguishes this approach from related fields, e.g. clustering \cite{FisherC,XuC,JainClust}, conceptual clustering \cite{FisherCC,michalski80} and multi-view clustering \cite{Bickel:2004, GambergerML14}. Redescription mining as an unsupervised technique is different from subgroup discovery \cite{wrobel1997SD,GambergerL02SDW,Lavrac:2004SDMLR,HerreraSD} and its ability to find bi-directional associations distinguishes it from association rule mining \cite{AgrawalAR,ZhangARM,HippARM} that provides uni-directional associations.

Existing redescription mining algorithms, cf. \cite{Ramakrishnan:2004,Zaki:2005,Parida:2005,GalloMM08,GalbrunV:2012,Mihelcic15LNAI,zinchenko2014masters} produce redescriptions using maximally two different views. This significantly limits the applicability of redescription mining to more complex problems. 

The main contribution of this work is a memory and time-complexity efficient framework for multi-view redescription mining, able to find redescriptions on datasets containing an arbitrary number of views. The methodology builds upon and extends our preliminary work \cite{MihelcicMW}, which demonstrated the feasibility of this approach. In addition to providing thorough results of an framework's performance and comparison to the naive way of creating multi-view redescription mining algorithms by utilizing existing two-view approaches, we provide experiments of using supplementing forest of Predictive Clustering trees (PCTs) \cite{MihelcicRF2017}, with \cite{BreskvarROS} (PCT-ROS) and without \cite{KocevSO} random output selection and using a supplementing forest of Extra multi-target Predictive Clustering trees \cite{KocevET} (EPCT) to increase the overall performance. Using a supplementing model obtained by training a Random Forest of PCTs was shown to increase accuracy and diversity of produced redescriptions when two views are used \cite{MihelcicRF2017}. In this work, we test the same hypothesis on datasets using more than two views and test supplementing models produced by training PCTs with random output selection and using random forest of Extra multi-target Predictive Clustering trees. We further test the feasibility of using Extra multi-target Predictive Clustering trees as a main rule-generating methodology, instead of using ordinary PCTs (which reduces the overall time complexity of the framework). As an additional technique that provides trade-off between accuracy and complexity, we test the framework's performance when using random projections on different subsets of pairs of initial views, instead of using all possible view pairs to create initial redescriptions (these that are later expanded to redescriptions containing queries constructed on all available views). Finally, we show examples of how this methodology can be used to increase the overall understanding of the outputs of different predictive machine learning models, which is an important contribution for the fields of machine learning and explainable data science.  

Section \ref{notation} contains explanations, notation, definitions and provides an illustrative redescription example, Section \ref{rw} contains the related work and motivates the use of techniques and models presented.  Section  \ref{methodology} provides a detailed description of the proposed framework for multi-view redescription mining, while Section \ref{complexity} contains the results of the complexity analyses of the framework. Section \ref{data} describes the data used to perform the experimental evaluation presented in Section \ref{evaluation}. This evaluation tests the ability of the proposed framework to discover new knowledge on three different datasets, compares its performance with naive implementation that uses existing two-view redescription mining approaches to create multi-view redescriptions and demonstrates the use of the proposed framework to increase the understanding of various machine learning models.  Finally, Section \ref{conclusions} contains conclusions and possible directions for future work.

\section{Notation and definitions}
\label{notation}
Input to the redescription mining algorithm consists of a set of entities ($E$) and a set of attributes ($V$), which are logically grouped in one or more views, denoted $W_i,\ i\geq 1$. Each variable $V_j,\ j\geq 1$ belongs to one of the views $W_i$. A redescription $R$ is a tuple of queries $R = (q_1,q_2,\dots,q_n)$, where each query $q_i$ describes a set of entities using only variables belonging to the corresponding view $W_i$ and all queries contained within a redescription must describe the same or very similar sets of entities. Similarity of these sets is measured by some relation, denoted as $\sim$. Queries are logical formulas built using variables and the logical operators of conjunction, disjunction and negation. These are the building blocks of a query language $Q$ (see \cite{GalbrunPhD} and \cite{MihelcicPhD} for more details). A formal definition of the redescription mining task from \cite{GalbrunPhD} is directly applicable in the multi-view setting:  

\begin{definition}
Given a set of entities $E$, a set of attributes $V$ describing these entities, a set of views $\mathcal{W}$, a query language $Q$, a similarity relation $\sim$ and a constraint set $\mathcal{C}$, the task of redescription mining is to find all redescriptions that satisfy the constraints in $\mathcal{C}$.
\end{definition}

\par 
The most commonly used similarity relation is the Jaccard index. Constraint set $\mathcal{C}$ includes conditions on the redescription support, the Jaccard index and the $p$-value (which we define in this section) but can also include constraints on the average redescription element and attribute Jaccard index (description and support redundancy) and complexity, defined as the normalized redescription query size (see Section \ref{evaluation} for formal definition). Even with a reasonable set of constraints, following the original definition of a task can potentially lead to creation of a large amount of patterns. Because of this, we aim to find a much smaller subset of patterns that satisfy the set of constraints $\mathcal{C}$ and optimizes broader set of measures, leading to representative and high quality subset that might be interesting to the end user.

The support set of a  query $q_i$ ($supp(q_i)$) is the set of all entities satisfying its conditions. The redescription $R=(q_1,q_2,\dots, q_n),\ n\in \mathbb{N}$ describes the entity $e$ if $e\in supp(q_i),\ \forall i\in \{1,2,\dots, n\}$. All entities described by a redescription compose a redescription support set ($supp(R)=supp(q_1)\cap supp(q_2)\cap \dots \cap supp(q_n)$). $R$ describes entities using $n$ queries built using attributes from $n$ different views, thus nViews$(R) = n$.
\par As an example, we present a redescription of a set of countries by using a trading view (view $1$), a population view (view $2$), a energy view (view $3$) and a country development and wealth view (view $4$). The redescription $R_{ex}=(q_{{ex}_1}, q_{{ex}_2},q_{{ex}_3},q_{{ex}_4})$ contains four queries. It is presented in Table \ref{tab:red1}. Variables of each query of the example redescription (e.g \texttt{E/I\_Cork\_Wood} - export to import contribution ratio of cork and wood, \texttt{E/I\_Road\_Vehicles} - export to import contribution ratio of road vehicles etc.), are connected with the conjunction ($\wedge$ - AND) operator. Numerical constraints denote the range of attribute values for entities contained in the redescription support set.

\begin{table*}[ht!]
\footnotesize
\caption{Example redescription $R_{ex}=(q_{{ex}_1}, q_{{ex}_2},q_{{ex}_3},q_{{ex}_4})$, with nViews$(R_{ex}) = 4$, that re-describes five countries: France, Germany, Italy, Japan and Spain.}
\label{tab:red1}
\begin{tabular}{l l}
$q_1:$ & $0.007\leq \text{\texttt{E/I\_Cork\_Wood}}\leq 1.305 \ \wedge\ 0.754 \leq\text{\texttt{E/I\_Road\_Vehicles}}\leq 8.098 \ \wedge\  
 0.958 \leq\text{\texttt{E/I\_Chemical\_Products}}\leq 1.631$ \\
$q_2:$ & $ 39.4\leq \text{\texttt{LABOR\_F}}\leq 53.5  \ \wedge\ 3.0 \leq\text{\texttt{MORT}}\leq 4.5 \ \wedge\ 8.27 \leq\text{\texttt{RUR\_POP}}\leq 31.42$\\
$q_3:$ & $3965.0 \leq \text{\texttt{ElectricityTotNetCapPPSol}}\leq 32643.0$\\
$q_4:$ & $1.396\cdot 10^{12}\leq \text{\texttt{GNIAtlas}}\leq 6.101\cdot 10^{12} \ \wedge\ -6.93\leq\text{\texttt{MON\_GROWTH}}\leq 6.605$
\end{tabular}
\end{table*}

Higher similarity among the sets of entities described by each of redescription's queries represents higher redescription accuracy. The Jaccard index quantifies this similarity and is used as a measure of redescription accuracy. It is defined as: 
\begin{equation}
\label{measures:js}
J(R) = \frac{|supp(q_1) \cap supp(q_2)\cap \dots \cap supp(q_n)|}{|supp(q_1) \cup supp(q_2)\cup \dots \cup supp(q_n)|}
\end{equation}

A statistical significance of some redescription $R=(q_1,q_2, \dots, q_n)$ is determined by testing the following null hypothesis: \emph{$supp(R)$ is obtained by joining randomly generated queries $q_1,\dots q_n$, where the probability of obtaining $q_i$ equals $|supp(q_i)|/|E|$}. The decision to accept or reject the null hypothesis is made by computing $p_{val}(R)$. This $p$-value represents the probability of obtaining a support set of a size equal to or larger than that of $supp(R)$, by combining $n$ randomly generated queries with marginal probabilities corresponding to the marginal probabilities of queries $q_1$, $q_2, \dots, q_n$. $p_{val}$ \cite{MihelcicMW} is computed from the binomial distribution: 
\begin{equation}
\label{measures:pval}
p_{val}(R)=\sum_{k=|supp(R)|}^{|E|} {|E|\choose k}(\prod_{i=1}^{n} p_i)^k\cdot(1-\prod_{i=1}^{n} p_i)^{|E|-k} 
\end{equation}
\noindent $|E|$ denotes the number of entities, $p_1 = |supp(q_1)|/|E|$, $p_2 = |supp(q_2)|/|E|, \dots, p_n = |supp(q_n)|/|E|$ are the marginal probabilities of obtaining $q_1$, $q_2, \dots, q_n$. 

Redescription presented in Table \ref{tab:red1} describes $5$ countries, has a Jaccard index value of $1.0$ and a $p$-value of $0.0$.

We use $Alg$ to denote an arbitrary multi-target regression (multi-label classification), machine learning algorithm. $\mathcal{M}$ denotes a model obtained after training the algorithm $Alg$ on some dataset. The \emph{rule-transformable} model denotes a model that can be transformed into a set of rules. The \emph{supplementing model} is a secondary (auxiliary) rule-transformable model used to create additional rules. These rules are used to increase accuracy and diversity of produced redescriptions and are discarded after redescription creation. 

$attrs(R)$ denotes a set of attributes used in redescription queries and $attr(R)$ the multi-set of all (potentially multiple) attribute occurrences in the queries of $R$. 

\section{Motivation and related work}
\label{rw}

Strong trends in different scientific domains encourage data collection in such a way that a set of entities or objects is measured, characterized or annotated from different contexts. This results in increased availability of large and complex datasets with multi-view aspects of objects.   

To gain better insight into some underlying phenomenon of interest, our first task is to identify  some regularities or correspondences that exist between these different aspects of objects. For example, one might want to characterize world countries through a correspondence between their demographic properties and economic trends. This is the motivating principle behind redescription mining \cite{Ramakrishnan:2004}, a data analysis task that aims at finding multiple characterizations of subsets of objects, where each subset is simultaneously characterized with descriptions constructed from different views.

 Introduced by Ramakrishnan et al. \cite{Ramakrishnan:2004}, who also proposed the first algorithm for obtaining redescriptions called CARTwheels, this task was initially performed using one view containing Boolean attributes. Such a setting was also adopted by Zaki \cite{Zaki:2005}, whose approach is based on a lattice of closed itemsets, and Parida \cite{Parida:2005}, who developed an approach based on a relaxation lattice. The problem with using only one view is that there is no way to make a logical separation between variables and explore associations between such sets of attributes. Early redescriptions contained a mix of attributes in both queries (which was also the maximal number of possible queries). Later work by Gallo et al. \cite{GalloMM08} introduced the logical separation of attributes into views, ensuring that each query contains variables only from the corresponding view. The greedy and MID algorithms \cite{GalloMM08}, based on frequent closed itemset mining, work by using maximally two views of Boolean attributes.  Galbrun and Miettinen \cite{GalbrunV:2012} extended the Greedy approach by Galo et al. \cite{GalloMM08}  to using Boolean, categorical and numerical attributes, with maximally two views. The same constraints on the number of views apply for the Split trees and Layered trees algorithms, developed by Zinchenko et al. \cite{zinchenko2014masters,zinchenkoConf} and the CLUS-RM algorithm, based on Predictive Clustering trees, developed by Mihelčić et al. \cite{Mihelcic15LNAI}.

 As can be seen, all state-of-the-art redescription mining paradigms are limited to mining redescriptions from two views, and this work is the first attempt to construct efficient multi-viewed redescription mining approach. The benefits expected from truly multi-viewed redescription mining approach are related to: (i) more accurate, complementary description of data, (ii) efficient pruning of patterns which leads to smaller execution times and more efficient memory consumption compared to the exhaustive approaches or naive extensions of the two-view redescription mining approaches to the multi-view setting, and (iii)  improved knowledge discovery  through possibility to discover higher order interactions and more rich explanatory capabilities.  For example, finding sets of river locations where there exist simultaneous co-habitation of distinct subsets of different plant and different subsets of animal species and describing these habitats by their chemical characteristics (a valid task in ecology research \cite{malmstrom10}) is very hard to obtain using current state of the art approaches. As developed, current approaches offer possibility to create redescriptions on pairs of views - currently there is no way to focus the search of these approaches to find explicitly only redescriptions with properties as described in the example above.  Redescriptions obtained on different pairs of views would mostly be mutually unrelated making it very difficult to obtain the required information. Naive extensions of these approaches to multi-view setting result in exponential increase of number of sub-queries and queries that need to be tested and combined into redescriptions, which makes this approach highly inefficient. A general multi-view redescription mining approach can also be utilized in the context of explainable machine learning and data science as it can be applied for  understanding of relations between different models and incorporating their
results in the data exploration process in an interpretable  manner. Here, multi-view extensions allow relating multiple models and explaining them and their relations with one or more distinct sets of original data attributes (not possible using only two views). When observing model-computed attribute importance for predicting some target concept, the general approach allows relating importance's obtained from multiple disjoint feature sets or relating importance's obtained from different models. Observing features deemed important (or predictive) by two or more different models increases confidence in the value of these features and can be used to prioritize potential experimental validation. 
 
 \subsection{The GCLUS-RM algorithm}
 
As a part of the related work, we present a slightly modified generalized version of the CLUS-RM algorithm \cite{mihelcic2017framework} which we call the GCLUS-RM. 
The generalised CLUS-RM algorithm (GCLUS-RM), presented in Algorithm \ref{alg:CLUSRM}, contains memory constraints on the maximal size of the redescription set and allows using an arbitrary, rule-transformable, model $\mathcal{M}$ obtained using some multi-target regression (multi-label classification), machine learning algorithm $Alg$ (lines $1$ to $10$). In this work $Alg$ equals the Predictive Clustering trees algorithm \cite{BlockeelPCT}.  PCTs are a generalization of Decision trees algorithm that are able to simultaneously predict multiple target variables. It uses variance reduction to determine the splits and uses information about cluster centroid which allows it to perform simultaneous clustering in the attribute and in the target space. 

Method \texttt{createInitialMs} takes as input the initial dataset (obtained by creating $|E|$ artificial examples by permuting the values of attributes of original examples, see \cite{Mihelcic15})  and trains the initial models using the algorithm $Alg$ to distinguish between the original and the artificial examples. Method \texttt{extractRulesFromM} transforms the obtained models into a set of rules and adds the non-redundant rules into rule-sets. It also marks which rules can be used for redescription construction (newly constructed and these from the previous iteration, see \cite{Mihelcic15}). \texttt{constructTargets} constructs target variables to be used by the algorithms in each algorithm iteration. Each rule produced on $W_i$ in the previous iteration is used as target to produce rules with similar support on $W_j$ (for more details see \cite{Mihelcic15}). Since the maximal size of a redescription set is limited, it is not allowed to have duplicate redescriptions in the set. This necessitates redundancy checks, which are performed when the conjunctive refinement procedure is used (see \cite{mihelcic2017framework}). Thus, this procedure (which iteratively improves redescription accuracy by joining its queries with queries of redescriptions whose support set is a superset or equal using a conjunction operator) is always included during redescription construction. The algorithm can be applied to an arbitrary pair of views from the set $MW=\{W_1,W_2,\dots,W_n\}$. In case maximal size of the redescription set is reached, the algorithm exchanges the newly produced redescription with the worst incomplete candidate from $\mathcal{R}$ (lines $11$ to $19$ in Algorithm \ref{alg:CLUSRM}). Method \texttt{createRedescriptions} (lines $12$ and $14$) combines marked rules into redescriptions using logical operators $\wedge,\ \vee$ and $\neg$. The method computes $\mathcal{O}(|\mathcal{R}|\cdot |r_{W_i}^{m}|\cdot |r_{W_j}^{m}|)$ set intersections to obtain redescriptions, where $r_{W_i}^{m},\ r_{W_j}^{m}$ are the marked subsets of rules.

\begin{algorithm*}[ht!]
\caption{The GCLUS-RM algorithm}\label{alg:CLUSRM}
\begin{algorithmic}[1]
\Require{First view data ($W_i$), Second view data ($W_j$), Constraints $\mathcal{C}$, Settings $\mathcal{S}$, Model generating algorithm $Alg$, Supplementing model generating algorithm $Alg'$}
\Ensure{A set of redescriptions $\mathcal{R}$}
\Procedure{GCLUS-RM}{}
\State  $[\mathcal{M}_{Wi_{init}}, \mathcal{M}_{Wj_{init}}]\leftarrow$ createInitialMs($W_i$, $W_j$, $Alg$)
\State $[r_{W_i},r_{W_j}]\leftarrow$ extractRulesFromM($\mathcal{M}_{Wi_{init}}, \mathcal{M}_{Wj_{init}}$)
\While{RunInd<$\mathcal{S}.$maxIter}
\State $[D_{W_i},D_{W_j}]\leftarrow$ constructTargets($r_{W_i}$,$r_{W_j}$)
\State $[\mathcal{M}_{W_i},\mathcal{M}_{W_j}]\leftarrow$ createMs($D_{W_i},D_{W_j}, Alg$)
\State $[r_{W_i},r_{W_j}]\leftarrow$extractRulesFromM($\mathcal{M}_{W_i},\mathcal{M}_{W_j}$)
\If{($\mathcal{C}.numSupplementModels>0$)}
\State $[\mathcal{M}_{W_i}',\mathcal{M}_{W_j}']\leftarrow$ createMs($D_{W_i},D_{W_j}, Alg'$)
\State $[r_{W_i},r_{W_j}] \leftarrow$ extractRulesFromM($\mathcal{M}_{W_i}',\mathcal{M}_{W_j}'$)
\EndIf
\If{($|\mathcal{R}|\leq \mathcal{C}.MaxExpansionSize$)}
\State $\mathcal{R} \leftarrow \mathcal{R} \cup  \text{createRedescriptions}(r_{W_i}, r_{W_j}, \ \mathcal{C})$
\Else
\State $\mathcal{R}'\leftarrow \text{createRedescriptions}(r_{W_i}, r_{W_j}, \ \mathcal{C})$
\For{($R\in \mathcal{R'}$)}
\State $R_{k}\leftarrow argmax_{R'\in \mathcal{R}} (J(R)-J(R')-(1-elemJ(R,R'))),\ J(R)>J(R')$, nViews($R'$)$<\mathcal{S}.n$
\State $\mathcal{R}\leftarrow\mathcal{R}\setminus R_{k} \cup R$
\EndFor
\EndIf
\If{($\mathcal{C}.numSupplementModels>0$)}
\State $[r_{W_i}, r_{W_j}]\leftarrow$removeSupplementRules($r_{W_i}, r_{W_j}$)
\EndIf
\EndWhile
\State \textbf{return} $\mathcal{R}$
\EndProcedure
\end{algorithmic}
\end{algorithm*}  
 Lines $8$-$10$ and $19$-$20$ demonstrate the use of supplementing rules derived from any rule-transformable supplementing model $\mathcal{M}'$ \cite{MihelcicRF2017}. In this work we use three models: a) The random forest of multi-target regression (multi-label classification) PCTs \cite{KocevSO}, b) The Random Forest of Extra randomized multi-target PCTs \cite{KocevET} and c) The Random Forest of multi-target regression PCTs with Random Output Selections \cite{BreskvarROS} (see Section \ref{subsec:motivation} for motivation and more details). Supplementing rules are removed from the rule sets using the \texttt{removeSupplementRules} method. This is done because these rules are only used to create and improve redescriptions and not to guide creation of new rules.   
 
All  multi-view approaches presented in this manuscript are based on the observation that for any redescription $R = (q_1,\dots, q_n)\in \mathcal{R}$ constructed using $n$ views, $J(R)\leq J(R^{*}_S)$ for $S\subseteq \{1,\dots,n\}$, where $R^{*}_S = (q_{1*}, q_{2*},\dots, q_{n*})$ and $q_{i*} = q_i,\ i\in S,\ q_{i*} = ?,\ i\notin S$. $q_{i*} = ?$ denotes that the query is missing, as a consequence redescription $R^{*}_S$ is incomplete - nViews($R^{*}_S$) $= |S|$. This gives a way to prune the redescription space. 
 
Following the guidelines above, our framework for multi-view redescription mining uses the GCLUS-RM algorithm to create satisfactory two-view redescriptions and then completes these redescriptions to $n$ views.

\section{A general framework for multi-view redescription mining}
\label{methodology}
In this section, we describe a generalized, memory-efficient framework for multi-view redescription mining that significantly extends the algorithm proposed in \cite{MihelcicMW}. As all general, previously developed algorithms for redescription mining, capable of working with different types of variables and missing values, this framework is also heuristic in nature. The framework is based on the generalized version of the CLUS-RM algorithm \cite{Mihelcic15LNAI} called GCLUS-RM introduced in Section \ref{rw} and the generalized redescription set construction procedure \cite{mihelcic2017framework} that allows creating multiple redescription sets of user-defined size, which satisfy various user-defined preferences.  

\subsection{Preliminaries}

In the continuation we present the notation used to shorten the pseudocode of the framework. \texttt{GCLUS\_RM}($W_i$, $W_j$, $\mathcal{C}$, $\mathcal{S}$ , $Alg$, $Alg'$) - denotes the execution of the two-view GCLUS-RM algorithm (see Algorithm \ref{alg:CLUSRM} and \cite{Mihelcic15LNAI} for the original CLUS-RM algorithm), with given view input parameters $W_i$ and $W_j$, the redescription constraint parameters $\mathcal{C}$, settings set $\mathcal{S}$ and the rule-transformable multi-target (multi-label classification) model generating algorithms $Alg$ and $Alg'$. The constraint set $\mathcal{C}$ includes the minimal redescription Jaccard index (redescription accuracy), maximal $p$-value (redescription significance, usually set to $0.01$), minimal (mostly $\geq 5$) and maximal redescription support size (usually $\leq 0.9\cdot |E|$) and the number of supplementing models (if a forest or an ensemble is used). Maximal support set size disables production of very general redescriptions or tautologies that are legal in the sense of the problem definition, but are usually not interesting. The constraint set also contains constraints on memory usage. The parameters \texttt{WorkSetSize} and \texttt{MaxRSSize} are user-defined parameters defining the maximum number of redescriptions that can be held in memory (\texttt{MaxRSSize}) and the maximum number of redescriptions to be held in memory between iterations and to be used to create the final set of redescriptions (\texttt{WorkSetSize}).  The model $\mathcal{M}$ obtained using algorithm $Alg$ is used to generate targets that connect two views and is transformed to rules used to create redescriptions (more detailed explanation can be seen in \cite{Mihelcic15LNAI}). Rules obtained from a model $\mathcal{M}'$, obtained using $Alg'$, are used to increase diversity and accuracy of produced redescriptions \cite{MihelcicRF2017}. In this work, we only test Predictive Clustering trees and Extra multi-target PCTs ($\mathcal{M}$) as models from which  main rules are generated. Using Extra multi-target PCTs as the model from which the main rules are generated has a significant implication on the time complexity of the GCLUS-RM algorithm and the framework for multi-view redescription mining (see Section \ref{complexity}). 

\subsection{High level overview}
\label{sec:HLO}
The proposed framework for multi-view redescription mining contains two important, mutually interleaving sets of algorithms:  
\begin{itemize}
    \item Algorithms for redescription and redescription set creation, which entail all algorithmic aspects required to create redescriptions, prevent the blow-up in number of created patterns and finally to obtain the output redescription sets.  
    \item Algorithms for memory management, which entail all algorithmic aspects required to minimize the overall memory usage of the framework and to provide a satisfactory trade-off between memory consumption and quality of the desired output.
\end{itemize}

\begin{figure*}[ht!]
\centerline{\includegraphics[width=0.7\textwidth]{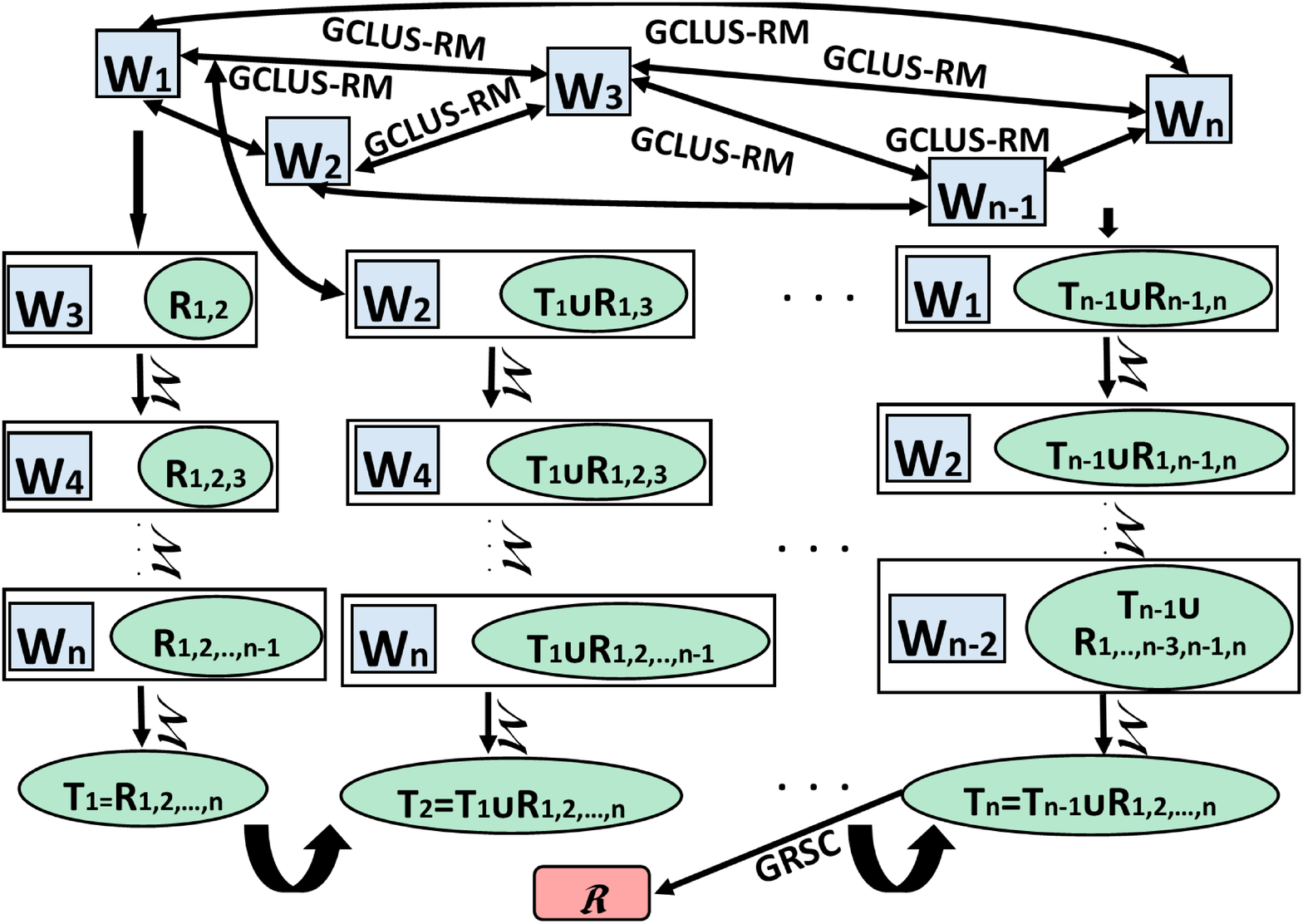}}
\caption{The framework uses a generalized version of the CLUS-RM algorithm (Section \ref{rw}, \cite{Mihelcic15LNAI}) to create two-view redescriptions on all pairs of views. Views are combined as denoted by numbers ($W_1$,$W_2$) first, ($W_1$,$W_3$) second, ($W_{n-1},W_n$) last. The produced redescriptions form targets used to construct an arbitrary rule-transformable multi-target  (multi-label) prediction model utilized to obtain corresponding rules on other views. Rule-producing models can be enhanced by using a Random Forest of arbitrary rule-transformable models as a supplementing model \cite{MihelcicRF2017} (we use PCTs with \cite{BreskvarROS} and without \cite{KocevSO} random output selections and the Extra multi-target PCTs \cite{KocevET}) . The final redescription set $T_n$ is used to create a set of redescription sets $\mathcal{R}$ using the generalized redescription set construction procedure (GRSC) \cite{mihelcic2017framework}.}
\label{fig:MWRM}
\end{figure*}

A high level overview explaining the set of procedures used to create redescriptions and the output redescription sets can be seen in Fig. \ref{fig:MWRM}. The main idea is to: a) iteratively select pairs of available views (until all pairs are exhausted), b) create two-view redescriptions using these views and the GCLUS-RM algorithm (Section \ref{rw}, \cite{Mihelcic15LNAI}), c) form targets from the available incomplete redescriptions and use them to train an arbitrary rule-transformable multi-target regression (multi-label classification) model on the consecutive available views , d) transform newly obtained models to rules and use them to complete the incomplete redescriptions, e) use the obtained complete redescriptions to create output redescription sets.

\begin{figure*}[h]
\centerline{\includegraphics[width=0.7\textwidth]{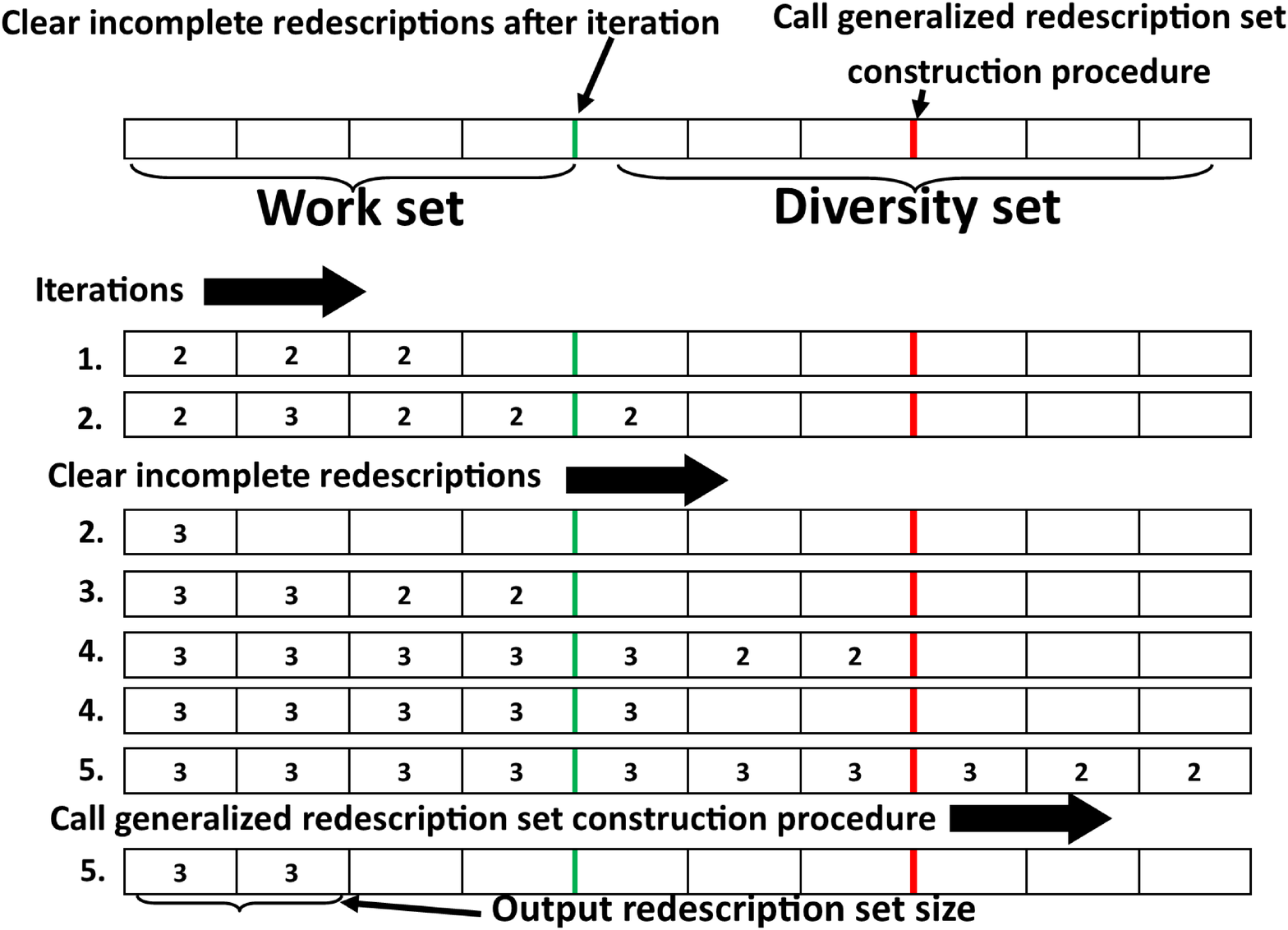}}
\caption{The memory model used in the framework for multi-view redescription mining. The available memory is divided in two initially empty parts: the work set and the diversity set. The example shows memory management during iterations on data containing three views.  After the number of redescriptions (complete and incomplete) in the redescription set exceeds the work set size, incomplete $2$-view redescriptions are discarded (after iteration $2$). Discarding of incomplete redescriptions continues until the number of redescriptions in the set is smaller or equal $t = (\mathcal{C}.MaxExpansionSize + \mathcal{C}.WorkSetSize)/2$ (the red mark). If the number of complete redescriptions exceeds $t$, the generalized redescription set construction procedure is called, selecting $r=2$ redescriptions (iteration $5$).}
\label{fig:MModel}
\end{figure*}

A high level overview of the memory management is depicted in Fig. \ref{fig:MModel}. The available memory consists of two parts, the \texttt{work set} and the \texttt{diversity set}. Sizes of these memory components are defined as the maximum number of redescriptions that they can store. Newly created redescriptions are first stored in the \texttt{work set} and in the \texttt{diversity set} only after \texttt{work set} memory is depleted. At the end of each iteration of the framework, all incomplete $2$-view redescriptions are cleared if the whole \texttt{work set} memory is full. There is a special threshold $t = (\mathcal{C}.MaxExpansionSize + \mathcal{C}.WorkSetSize)/2$ where $\mathcal{C}.MaxExpansionSize$ is the total amount of memory that can be used to store redescriptions (the red line in Fig. \ref{fig:MModel}). If the number of stored redescriptions crosses this number, incomplete redescriptions are discarded starting from $2$-view redescriptions, $3$-view redescriptions etc. until memory consumption is smaller than this threshold. The logic is to keep redescriptions that are very close to being complete as long as possible in memory. However, if the amount of complete redescriptions stored in memory is larger than $t$, the generalized redescription set construction procedure \cite{mihelcic2017framework} is called to reduce the number of redescriptions contained in memory to the predefined output set size. 

\subsection{Description of the framework for multi-view redescription mining}

This section contains analyses of two important algorithms (Algorithm \ref{alg:algorithm} and Algorithm \ref{alg:compReds}) constituting the proposed framework for multi-view redescription mining. In the continuation of the text, all line numbers refer to the corresponding lines of Algorithm \ref{alg:algorithm} until the description of the procedure \texttt{completeRedescriptions}, where all line numbers refer to the corresponding lines of Algorithm \ref{alg:compReds}.

\subsubsection{Redescription and redescription set creation}
The general framework for multi-view redescription mining (see Algorithm \ref{alg:algorithm}) is run over $\mathcal{S}.NRndRest$ random initializations of input dataset (lines $3-4$, see also explanation of a method \texttt{createInitialMs}). The GCLUS-RM algorithm is used to produce two-view redescriptions on each pair of available views (lines $5-9$). The key step that allows efficient multi-view redescription mining is to use these incomplete redescriptions as targets to produce matching rules on remaining views. Matching rules are obtained using any multi-target regression or multi-label classification algorithm able to produce rule-transformable models $\mathcal{M}_k$ and $\mathcal{M}_k'$ (lines $11$-$13$).

The described procedure allows solving the task by computing $\mathcal{O}({n \choose 2}\cdot |\mathcal{R}|\cdot |r_{W_i}^{m}|\cdot |r_{W_j}^{m}|)$ set intersections instead of $\mathcal{O}(\mathcal{R}\cdot \prod_{i=1}^{n} |r_{W_i}^{m}|)$ that would be needed if rules were computed on each view and then combined into multi-view redescriptions. Naive extension of the CLUS-RM algorithm requires either computing many rules using pairwise CLUS-RM (e.g compute rules on $W_1$ using rules from $W_2$ as targets, then computing rules on $W_3$ using rules from $W_2$ as targets etc. and than exhaustively combining these rules - which is time and memory consuming) or adding rules from all views as targets at each step of the algorithm. This still does not guarantee obtaining rules satisfying constraints from all views but can cause significant technical problems. Very large number of simultaneous targets is hard to satisfy, thus attempts to use it usually result in inaccurate models. 

\begin{algorithm*}[ht!]
\caption{A general framework for multi-view redescription mining}\label{alg:algorithm}
\begin{algorithmic}[1]
\Require{Available views $MW=\{W_1,\dots,W_n\}$, Constraints $\mathcal{C}$, Settings $\mathcal{S}$, Model generating algorithm $Alg$, Supplementing model generating algorithm $Alg'$}
\Ensure{A set of reduced redescription sets $\mathcal{R}$}
\Procedure{MW-RM}{}
\State $\mathcal{R}_{all}\leftarrow \emptyset$
\For{(nrand = 0; nrand<$\mathcal{S}.NRndRest$; nrand++)}
\State $MW'=\{W_1',\dots,W_n'\}\leftarrow$\texttt{initializeViews()}
\For{(i=0; i<$|MW|-1$; i++)}
\For{(j=i+1; j<$|MW|$; j++)}
\State RunInd$\leftarrow 0$
\State $\mathcal{R}_{i,j}\leftarrow$ \texttt{GCLUS\_RM}($W_i'$, $W_j'$, $\mathcal{C}$, $Alg$, $Alg'$, $\mathcal{S}$)
\State $R_{all}\leftarrow R_{all}\cup R_{i,j}$
\For{(k=0, $k\notin \{i,j\}$; k<$|MW|$; k++)}
\State $DW_k\leftarrow$\texttt{constructTargets}($W_k$, $\mathcal{R}_{all}$)
\State $\mathcal{M}_k\leftarrow$ createMs$(DW_k,\ Alg)$
\State $r_k \leftarrow$\texttt{extractRulesFromM($\mathcal{M}_k$)}
\If{($\mathcal{C}.numSupplementModels>0$)}
\State  $\mathcal{M}'_k\leftarrow$ createSupplementingMs$(DW_k,\ Alg')$
\State  $r_k\leftarrow$\texttt{extractRulesFromM($\mathcal{M}'_k$)}
\EndIf
\State $\mathcal{R}_{all}\leftarrow$\texttt{completeRedescriptions}($\mathcal{R}_{all},r_k$, $\mathcal{S}.Op$, $\mathcal{C}$, $W_k'$)
\If{($\mathcal{C}.numSupplementModels>0$)}
\State $r_k\leftarrow$removeSupplementRules($r_k$)
\EndIf
\State $\mathcal{R}_{all}\leftarrow$\texttt{normalizeMemory}($\mathcal{R}_{all}$, $\mathcal{C}$, $\mathcal{S}$)
\EndFor
\EndFor
\EndFor
\EndFor
\For{($nws = 2;\ nws<|MW|; nws++$)}
\State  $\mathcal{R}_{all}\leftarrow$\texttt{removeIncomplete}($\mathcal{R}_{all}$, $nws$)
\EndFor
\State $\mathcal{R}_{all}\leftarrow$\texttt{minimizeQueries}($\mathcal{R}_{all}$)
\State $\mathcal{R}_S\leftarrow$\texttt{GRSC}($\mathcal{R}_{all}$, $\mathcal{S}.\mathcal{W}$, $\mathcal{S}.r$)
\State \textbf{return} $\mathcal{R}_S$
\EndProcedure
\end{algorithmic}
\end{algorithm*}


 The function \texttt{constructTargets} (line $11$) works similarly as the target construction procedure defined in \cite{Mihelcic15LNAI}. Each incomplete redescription in the set $\mathcal{R}_{all}$ constitutes one target variable in a newly constructed multi-target regression (classification) task. Every entity redescribed by some redescription $R_k\in \mathcal{R}_{all}$ has a value $1.0$ for the k-th target variable. If an entity is not redescribed by a redescription $R_k$ it has a value $0.0$ for this variable. A multi-target regression or multi-label classification rule-transformable model ($\mathcal{M}_k$ and potentially $\mathcal{M}_k'$) are trained on a dataset containing attributes of the $k$-th view and the aforementioned target variables, to construct rules which are used to complete redescriptions. Rules obtained from the supplementing model (lines $14$-$16$) are used to increase the diversity and the accuracy of the produced redescriptions \cite{MihelcicRF2017}. The newly produced rules are used to complete potentially incomplete redescriptions (line $17$ ). Following the established procedure of using supplementing models  \cite{MihelcicRF2017}, rules obtained from the supplementing model are discarded after redescription creation (lines $18-19$).  The fast growth of the number of produced redescriptions is controlled with the two earlier explained parameters, \texttt{MaxExpansionSize} and \texttt{WorkSetSize} inside method \texttt{normalizeMemory} (line $20$). This procedure is described in Section \ref{sec:HLO} and Figure \ref{fig:MModel}. All incomplete redescriptions are discarded before the creation of the final result sets (lines $21$ and $22$). The redescription query size is reduced by using the query size minimization procedure introduced in \cite{Mihelcic15LNAI} (line $23$). Finally, the generalized redescription set construction procedure (GRSC) is used to create a resulting set of reduced redescription sets (line $24$). \texttt{GRSC}($\mathcal{R}$,$\mathcal{W}$,$r$)  \cite{mihelcic2017framework}  takes as input a redescription set $\mathcal{R}$, a user-defined quality measure importance weight matrix $\mathcal{W}$ and an integer $r$ denoting the required size of output redescription set. The procedure returns one or more optimized redescription sets of size equal to or smaller than $r$. The quality measure importance weight matrix allows users to influence the structure of the resulting redescription sets by giving higher emphasis to a subset of redescription quality measures (e.g giving higher weight to redescription Jaccard index will cause the procedure to favour redescription accuracy over diversity or reduced complexity). By default, $\mathcal{W}$ is a $1\times 5$ matrix assigning equal weight $\frac{1}{5}$ to each quality measure (see Section \ref{evaluation} for a list of measures and their definition). 

\subsubsection{Completing incomplete redescriptions}
The procedure \texttt{completeRedescriptions} from Algorithm \ref{alg:compReds} iterates over all incomplete redescriptions (line $2$) and attempts to complete them with some rule from rule set $r$ (line $3$). If adding a new query to some existing incomplete redescription $R$ satisfies the accuracy constraints, a new redescription is created (lines $4$-$5$). Since the \texttt{conjunctiveRefinement} procedure is used (a procedure that can increase redescription accuracy using existing redescriptions with support set equal to a target redescription or a superset thereof, see \cite{mihelcic2017framework}), a redescription is added to the set of redescriptions only if there is no redescription with equal support and maximal accuracy (line $6$).  If the newly created redescription has the required accuracy and support set size characteristics, it is non-redundant and there is sufficient amount of memory available, it is added to the set of all redescriptions through the \texttt{addDiscardOrReplace} method (line $10$). It the available memory is full, the newly produced redescription aims to replace the most similar redescription with the largest difference in accuracy. If such candidate can not be found, newly produced redescription is discarded.

\begin{algorithm*}[ht!]
\caption{Complete redescriptions}\label{alg:compReds}
\begin{algorithmic}[1]
\Require{Redescriptions $\mathcal{R}$, Rules $r$, Operators $op$, Constraints $\mathcal{C}$, View $W_i$}
\Ensure{A set of redescriptions $\mathcal{R}$}
\Procedure{completeRedescriptions}{}
\For{($R\in \mathcal{R},\ R.q_i = \emptyset$)}
\For{($r_j \in r$)}
\If{(J($supp(R)\cap supp(r_j), \cup_{q_i \in R} supp(q_i) \cup supp(r_j)$)$\geq \mathcal{C}.minJS$)}
\State $R_{new}\leftarrow R.insertQuery(r_j,W_i)$
\State $R_{new}\leftarrow $\texttt{conjunctiveRefinement($R_{new},\mathcal{R}$)}
\If{$R_{new} = \emptyset$}
    \State continue
\EndIf 
\If{($supp(R_{new})\in [\mathcal{C}.minSupp,\mathcal{C}.maxSupp]\ \wedge \ J(R_{new})\geq \mathcal{C}.minJS$)}
\State $\mathcal{R}\leftarrow$\texttt{addDiscardOrReplace($R_{new},\ \mathcal{R}$)}
\EndIf
\EndIf
\If{($\neg \in op\ \wedge J(supp(R)\cap supp(\neg r_j), \cup_{q_i \in R} supp(q_i) \cup supp(\neg r_j))\geq \mathcal{C}.minJS$)}
\State $R_{new}\leftarrow R.insertQuery(\neg r_j,W_i)$
\State $\mathcal{R}\leftarrow$\texttt{addDiscardOrReplace($R_{new},\ \mathcal{R}$)}
\EndIf
\EndFor
\EndFor
\If{($\not\exists R\in \mathcal{R},\ nViews(R)<n\ \wedge \ \mathcal{C}.MaxExpansionSize = |\mathcal{R}|$)}
\State $\mathcal{R}\leftarrow $\texttt{refineByQueryReplacement($\mathcal{R}\setminus$\texttt(used($\mathcal{R}$)),$\ r\setminus$\texttt(used($r$)))}
\EndIf
\If{$\vee \in op$}
\State $\mathcal{R}\leftarrow$\texttt{refineDisjunctive($\mathcal{R},\ r,\ W_i$)}
\EndIf
\State \textbf{return} $\mathcal{R}$
\EndProcedure
\end{algorithmic}
\end{algorithm*}  

\noindent Lines $11-13$ demonstrate the use of negation operator. If there are no incomplete redescriptions remaining in the redescription set after completing lines $2$ to $13$ and there is no available memory, all redescriptions not tested or completed in lines $2$ to $13$  are refined using remaining queries (these not used in lines $2$ to $13$, obtained with \texttt{used}() function). This is done by exchanging their query (for the corresponding view) with some rule from $r$ or it's negation if this increases redescription accuracy (lines $14$ and $15$). Use of the refinement by query replacement is limited to preserve overall query diversity. Method \texttt{refineDisjunctive}  (line $17$) takes as input a redescription set $\mathcal{R}$, a rule set $r$, a given view $W_i$ and it tries to improve the accuracy of each redescription using rules from a set $r$ constructed on view $W_i$. Some rule $r_j\in r$ is used as a disjunctive refinement rule of a redescription $R_p\in \mathcal{R}$ if it maximizes: J($\cap_{q_k \in R_p, k\neq i} supp(q_k)\setminus supp(R_p)$, $supp(r_j)$) or if its negation maximizes: J($\cap_{q_k \in R_p, k\neq i} supp(q_k)\setminus supp(R_p)$, $supp(\neg r_j)$) and the newly obtained redescription $R_j'$ ( with $R'.q_{W_i}\leftarrow R'.q_{W_i}\vee r_{indMax1}$ or $R'.q_{W_i}\leftarrow R'.q_{W_i}\vee \neg r_{indMax1}$) satisfies constraints $\mathcal{C}$. Finally, the extended set of redescriptions is returned in line $18$. Adhering to memory constraints, this procedure potentially extends incomplete redescription throughout multiple iterations, creating diverse set of candidates which reduces the effects of seemingly greedy updates.

\subsection{Motivation for using additional tree-based models}
\label{subsec:motivation}
Rule-transformable models differ by the characteristics and the number of obtainable rules. Understanding the effects of utilizing models obtained from different machine learning algorithms for redescription construction is an important research direction. We focus on models with very interesting properties obtained using: the Extra multi-target PCTs \cite{KocevET} and the PCTs with random output selection \cite{BreskvarROS}. 

\subsubsection{Extra multi-target Predictive Clustering trees}
Extra multi-target PCTs \cite{KocevET} introduce random split selection into the construction of Predictive Clustering trees for multi-target prediction  \cite{KocevSO}. For each split, the extra multi-target PCTs select $k$ random attribute-value pairs and then compute the best candidate using the variance reduction. The same measure is used in the regular PCTs to determine the best split, however PCTs test all possible attribute-value pairs.  Because of the high randomization, the extra multi-target PCTs are weak models, but they have been shown to work well in the ensemble setting \cite{KocevET}. The main advantage of the Extra multi-target PCT approach is the reduced time complexity compared to the regular PCTs. 

In this work, we investigate using the Extra multi-target PCTs for creating redescriptions. There are two potential benefits of using this algorithm: a)  due to high level of randomization, Extra multi-target PCTs may produce different redescriptions from PCTs (increasing diversity), b) when used as a generating model, the Extra multi-target PCTs decrease the overall time complexity of the framework for multi-view redescription mining (see Section \ref{complexity}). There are also some drawbacks, the main is caused by the large width of the Extra multi-target PCTs (also discussed in \cite{KocevET}), which potentially creates a higher number of rule-pairs to be tested as compared to regular PCTs or ensembles thereof. Thus, the development of efficient approaches for rule selection to obtain accurate redescriptions is of high priority. 

\subsubsection{Predictive clustering trees with random output selections}
The ensembles of Predictive Clustering trees with random output selections \cite{BreskvarROS} train each PCT on a randomly selected subset of the target labels. Such procedure has been show to be able to outperform the ensemble of regular PCTs on several different datasets \cite{BreskvarROS}. 

Using a methodology that can utilize subsets of target labels is especially important in the step of the proposed multi-view redescription mining framework tasked with the completion of the incomplete redescriptions. Completion is done by using redescriptions as targets and training multi-target regression (multi-label classification) model to produce rules that allow accurate completion of these incomplete redescriptions. Since redescriptions do not generally have a strictly hierarchical structure, but are overlapping (many pairs have even disjoint support sets), multi-target regression models trying to simultaneously satisfy all targets mostly fail to do so for a subset of targets.

Using ensembles of Predictive Clustering trees with random output selections may increase the overall accuracy of the framework. The main disadvantage of this approach is a slightly higher execution time due to  sampling of a target space (although it does not affect the overall computational complexity of the approach, see Section \ref{complexity}). 

\section{Computational complexity}
\label{complexity}

We present the worst-case and the average-case time complexity analyses for the framework capable of performing multi-view redescription mining with the use of multi-target PCTs \cite{KocevSO}, multi-target PCT-ROS \cite{BreskvarROS} and the EPCTs \cite{KocevET} as algorithms used to create the supplementing models and using PCTs and EPCTs as algorithms that produce the main rule generating models (these models are transformed to rules used to explore the search space).

\subsection{Complexity using PCTs to create the main rule-generating model}
The time complexity of creating a Predictive Clustering tree model is $\mathcal{O}(m\cdot |E|\cdot log_2^2(|E|)) + \mathcal{O}(S\cdot m\cdot |E|log_2|E|) + \mathcal{O}(|E|log_2(|E|))$ (see \cite{KocevET}), where $m$ denotes the number of attributes, $|E|$ the number of entities contained in the data and $S$ the number of target variables. The number of target variables is constrained in our framework with $|\mathcal{R}|$, number of rules $z$ or user defined parameter \texttt{NumTarget}$<|\mathcal{R}|$. In case user-defined parameter is used, PCT is trained $\lfloor \frac{z'}{NumTarget} \rfloor +1$ times (where $z'$ represents the total number of targets - rules or redescriptions). Taking this into account, the average case time complexity of a GCLUS-RM algorithm, using PCTs as generating model and the conjunctive refinement procedure is 
$\mathcal{O}((|V_1|+|V_2|)\cdot |E|\cdot log_2^2(|E|) + S\cdot (|V_1|+|V_2|)\cdot |E|\cdot log_2|E| + z^3\cdot |E|)$, where $|V_i|$ denotes the number of attributes in the $i$-th view. The worst case time complexity (given inadequate hashing function) equals $\mathcal{O}((|V_1|+|V_2|)\cdot |E|\cdot log_2^2(|E|) + S\cdot (|V_1|+|V_2|)\cdot |E|\cdot log_2|E| + z^3\cdot |E|^2)$. The time complexity of the GCLUS-RM dominates the time complexity of the PCT. It is also worth noting that in between GCLUS-RM executions, maximal size of a redescription set is constrained and can be considered a constant. Thus, the time complexity of a generalized redescription set construction procedure equals $\mathcal{O}(|E|)$. Using this, the average case time complexity of the algorithm for multi-view redescription mining, using GCLUS-RM with PCT generating model is $\mathcal{O}(\sum_{i=1}^{n-1}\sum_{j=i+1}^{n}(
(|V_i|+|V_j|)\cdot |E|\cdot log_2^2(|E|) + S\cdot (|V_i|+|V_j|)\cdot |E|\cdot log_2|E| + z^3\cdot |E|)$, which is $\mathcal{O}((n-1)\cdot ((\sum_{i=1}^{n} |V_i|)\cdot |E|\cdot log_2^2(|E|) + S\cdot (\sum_{i=1}^{n} |V_i|)\cdot |E|\cdot log_2|E|)+$  $(n\cdot(n-1)/2)\cdot z^3\cdot |E|)$, where $n$ denotes the number of views. Since $n<<min{|V_i|,|E|},i\leq n$ and $n^2<<|E|$ in most real applications it can be considered a constant. Thus, the average case time complexity of the algorithm is $\mathcal{O}((\sum_{i=1}^{n} |V_i|)\cdot |E|\cdot log_2^2(|E|) + S\cdot (\sum_{i=1}^{n} |V_i|)\cdot |E|\cdot log_2|E|)+ z^3\cdot |E|)$. Similarly, the worst case time complexity of the framework equals: $\mathcal{O}((\sum_{i=1}^{n} |V_i|)\cdot |E|\cdot log_2^2(|E|) + S\cdot (\sum_{i=1}^{n} |V_i|)\cdot |E|\cdot log_2|E|)+ z^3\cdot |E|^2)$.

\subsection{Complexity using EPCTs to create the main rule-generating model}
\label{sub:cet}
The time complexity of creating the Extra multi-target PCTs is $\mathcal{O}(k\cdot S\cdot log_2|E|) + \mathcal{O}(|E|\cdot log_2(|E|))$ \cite{KocevET}, where $k$ denotes the number of randomly selected attribute splits that are evaluated to determine the best candidate. Given this, the average case time complexity of the GCLUS-RM is $\mathcal{O}(k\cdot S\cdot log_2(|E|) + |E|\cdot log_2|E| + z^3\cdot |E|)$, or the worst case time complexity $\mathcal{O}(k\cdot S\cdot log_2(|E|) + |E|\cdot log_2|E| + z^3\cdot |E|^2)$. Despite having much lower complexity compared to the GCLUS-RM with PCT generating model, due to the bushiness and width of the Extra multi-target PCTs (as described in \cite{KocevET}), using this model causes significant increase in the $z$  constant compared to using regular PCTs. The overall complexity of the framework for multi-view redescription mining, using Extra multi-target PCTs as rule-generating model has average time complexity:  $\mathcal{O}((n\cdot(n-1)/2)\cdot( k\cdot S\cdot log_2(|E|) + |E|\cdot log_2|E|+ z^3\cdot |E|))$. Given $n<<min{|V_i|,|E|},i\leq n$ and $n^2<<|E|$, the overall complexity is identical to the complexity of the GCLUS-RM algorithm using the Extra multi-target PCTs, $\mathcal{O}(k\cdot S\cdot log_2(|E|) + |E|\cdot log_2|E|+ z^3\cdot |E|)$. 

\subsection{Complexity of using Random Forest of supplementing models}

We use three types of supplementing models: a) Random Subspaces of Predictive Clustering trees, b) Random Forest of Extra tree multi-target models and c) Random Forest of Predictive Clustering trees with output selections. When Random Subspaces of Predictive Clustering trees are learned, first a random attribute subset of size $as = max(\lceil |V_i|\cdot (1-\sqrt[z]{(1-p)}) \rceil, log_2(|V_i|))$ is selected \cite{MihelcicRF2017}, where $p$ equals the desired probability of obtaining an attribute in a split of every tree in a forest. Next, the regular PCT model is learned on each attribute subset. Thus, the overall complexity of building such a Forest is: $\mathcal{O}(as\cdot |E|\cdot log_2^2(|E|)) + \mathcal{O}(S\cdot as\cdot |E|log_2|E|) + \mathcal{O}(|E|log_2(|E|))$. Learning a Random Forest of Extra PCTs has equal complexity $\mathcal{O}(k\cdot S\cdot log_2(|E|) + |E|\cdot log_2|E|+ z^3\cdot |E|)$, but here $k\leq as$. Learning a Random Forest of Predictive Clustering trees with output selections has a complexity of $\mathcal{O}(as\cdot |E'|\cdot log_2^2(|E'|)) + \mathcal{O}(S\cdot as\cdot |E'|log_2|E'|) + \mathcal{O}(|E'|log_2(|E'|))+ \mathcal{O}(|E|)$, where $E'$ denotes the fraction of the input training data used in Bagging (usually, $|E'| = 0.632\cdot |E|$ (see \cite{KocevSO}). Thus, learning PCT generating models has higher time complexity than learning supplementing models. Using supplementing models does not increase overall complexity when PCTs are used as a rule generating model. When Extra multi-target PCT algorithm is used as a generating model, using Forest of PCTs as supplementing model increases the time complexity of the approach but no more than the time complexity of using PCTs as generating model. The main problem in using large forests is potentially large increase in the constant $z$.

\section{Data description}
\label{data}

We use three different datasets to evaluate the proposed multi-view redescription mining methodology: a) the Country dataset, b) the River water quality dataset and c) the Phenotype dataset. The last two datasets are also used in a use case depicting the application of the proposed methodology to increase the understanding of machine learning predictive models, to help in model selection or in construction of a ensemble of machine learning models.

\begin{itemize}
\item[a)] The Country dataset contains $141$ entities (world countries) which are described with $4$ different views. The information is about the countries for the year $2012$. Country trade, consisting of $309$ numerical attributes and obtained from the UNCTAD database \cite{UNCTAD}  makes the first view. The second view, describing the population of these countries, consists of $21$ numerical attributes. Part of this data was obtained from the World bank \cite{wb} and part from the UN \cite{UN} database. The third view contains $47$ numerical attributes describing energy production and consumption of these countries. The fourth view describes different aspects of country development and wealth (agriculture, work, financial and ecological indicators) with $33$ numerical attributes. The data contained in the third and the fourth view was obtained from the UN database \cite{UN}. All views have missing values.
\item[b)] The River water quality dataset (Slovenian Water) \cite{Dzeroski2000WC} contains $3$ views describing $1061$ water samples taken from the Slovenian rivers by the Hydrometeorological Institute of Slovenia. The first view contains $16$ numerical attributes representing physical and chemical measurements of water quality (e.g. biological oxygen demand, chlorine concentration, etc). The second view contains the occurrence frequency of $7$ different plant species, whereas the third view contains occurrence frequency of $7$ different animal species. The frequencies are coded as: $0$-not present, $1$-incidental occurrence, $3$-frequently occurring and $5$-abundantly occurring.
\item[c)] The phenotype dataset \cite{BrbicLandscape} has three views, all containing numerical attributes with missing values. Attributes contained in these views describe $92$ entities (phenotypic properties of different microbial species). All $811$ attributes contained in the first view are positive feature importance scores. Features are metagenomic co-occurrences (co-occurrence of species across environmental sequencing data sets) used by the Random Forest algorithm to predict the presence of these phenotypic properties in different microbial organisms. Similarly, the second view contains $419$ attributes that represent the feature ranking scores, where the features represent proteome composition - relative frequencies of amino acids. The third view contains $990$ attributes that represent feature ranking scores, where features represent genomic signatures of translation efficiency in gene families.
\end{itemize}

\section{Experiments and results}
\label{evaluation}


In this section we present the naive implementation of multi-view redescription mining algorithm using existing $2$-view redescription mining approaches. We define additional redescription and redescription set evaluation measures required to provide full information about method performance and present the evaluation results of the proposed framework for multi-view redescription mining. 

\subsection{Naive algorithm for multi-view redescription mining}
\label{naive}


The naive implementation of a multi-view redescription mining algorithm includes mining $2$-view redescriptions on all pairs of available views and combining these incomplete redescriptions into redescriptions containing all available views. Definition of two operators is required to construct the pseudocode of the algorithm for naive multi-view redescription mining.
 
\par \vspace{2mm}\noindent
Given two incomplete redescriptions $R^{*}_{1, S_1}$ and $R^{*}_{2, S_2}$, where $S_1,\ S_2 \neq \emptyset$ we define the operation: \[   
R^{*}_{1,S_1} \oplus R^{*}_{2,S_2} = 
     \begin{cases}
       \{R^{*}_{new, S'}\},\ S' = S_1 \cup S_2,\  S_1 \cap S_2 = \emptyset,\\
      q_{new,i*} = q_{j,i*}, i\in S_j, j\in\{1,2\}\vspace{5mm}\noindent\\ 
       \{R^{*}_{new_1, S'},\ R^{*}_{new_2, S'}\},\ S' = (S_1 \cup S_2), \\ \hspace{35mm} k = argmin_s\\
       \hspace{32mm} s\in (S_1\cap S_2) \neq \emptyset \\
       \hspace{35mm} S_1\nsupseteq S_2, \\
       q_{new_1,i*} = q_{1,i*}, i\in S_1, \\
       q_{new_1,i*} = q_{2,i*}, i\in S_2\setminus \{k\},\\
       q_{new_2,i*} = q_{1,i*}, i\in S_1\setminus \{k\}, \\
       q_{new_2,i*} = q_{2,i*}, i\in S_2\vspace{5mm}\noindent\\
	 \emptyset,\   S' = S_1 \cup S_2,\  S_1 \supseteq S_2

     \end{cases}
\]

\par \vspace{2mm}\noindent
Given two sets of incomplete redescriptions: $\mathcal{R}_1$ containing views $S_1\neq \emptyset$ and $\mathcal{R}_2$ containing views $S_2\neq \emptyset$, we define: 

\noindent $\mathcal{R}_1 \otimes \mathcal{R}_2 = \{R^{*}_{1,S_1},\ R^{*}_{1,S_1}\in \mathcal{R}_1\}\cup \{R^*_{1,S_1}\oplus R^*_{2,S_2},\ R^*_{1,S_1}\in \mathcal{R}_1,\ R^*_{2,S_2}\in \mathcal{R}_2\}$.

The naive algorithm for multi-view redescription mining is presented in Algorithm \ref{alg:naiveMWalgo}. The algorithm uses any $2$-view redescription mining algorithms to construct incomplete redescriptions containing two queries (lines $5$, $6$ and $7$) in Algorithm \ref{alg:naiveMWalgo}. The resulting two-view redescription sets are denoted $R^*_{k,\{i,j\}}$, where $k\leq {n\choose 2}$ denotes the index of a set and $i,\ j\leq n,\ i<j$ denote indices of views used to create incomplete redescriptions. The algorithm creates incomplete sets sequentially in order $R^*_{1,\{1,2\}},\ R^*_{2,\{1,3\}},\dots R^*_{n,\{1,n\}},\dots,\ R^*_{{n\choose 2},\{n-1,n\}}$. These incomplete sets are combined in multi-view redescriptions using earlier defined operator of incomplete redescription set joining ($\otimes$) in lines $8$ and $9$. This operator combines only those redescriptions so that the number of views of the resulting redescription is larger than the number of views of the initial redescriptions (notice that the second operand always contains exactly two views). When a pair of incomplete redescriptions contains one query from the same view, there are two ways to join them to increase the number of views and they are both explored by the algorithm. Since the resulting set contains many incomplete redescriptions, these are first filtered out (line $10$). Redundant redescriptions are filtered out (those describing very similar entities,  having entity Jaccard index $>$ than a predefined threshold $perc$, as some other more accurate redescription contained in the set). Finally, a redescription set containing complete multi-view redescriptions is returned to the user.

\begin{algorithm*}[ht!]
\caption{Naive multi-view redescription mining}\label{alg:naiveMWalgo}
\begin{algorithmic}[1]
\Require{Available views $MW=\{W_1,\dots,W_n\}$, Constraints $\mathcal{C}$, Settings $\mathcal{S}$, Two-view RM algorithm $AlgRM$}
\Ensure{A set of reescriptions $\mathcal{R}$}
\Procedure{MW-RMNaive}{}
\State $\mathcal{R}_{incomplete}\leftarrow \emptyset$
\State $\mathcal{R}_{all}\leftarrow \emptyset$
\State $\mathcal{R}\leftarrow \emptyset$
\For{($(W_i,\ W_j)\in MW,\ i,j = 1\dots n,\ i < j$)}
\State $\mathcal{R}^{*}_{k,\{i,j\}}\leftarrow AlgRM(W_i,W_j, \mathcal{C}, \mathcal{S}),\ k\leq {n\choose 2}$
\State $\mathcal{R}_{incomplete}\leftarrow \mathcal{R}_{incomplete}\cup \{\mathcal{R}^{*}_{k,\{i,j\}}\}$ 
\EndFor

\For{($(\mathcal{R}^{*}_{i, S_i},\ \mathcal{R}^{*}_{j,S_j})\in \mathcal{R}_{incomplete},\ i,j = 1 \dots {n\choose 2},\ i< j$)}
\State $\mathcal{R}_{all}\leftarrow \mathcal{R}_{all}\cup (\mathcal{R}^{*}_{i, S_i} \otimes \mathcal{R}^{*}_{j,S_j})$
\EndFor

\State $\mathcal{R}_{all}\leftarrow $\texttt{filterIncomplete}($\mathcal{R}_{all}$)
\State $\mathcal{R}\leftarrow $\texttt{filter($\mathcal{R}_{all}, \mathcal{S}.perc$)}

\State \textbf{return} $\mathcal{R}$
\EndProcedure
\end{algorithmic}
\end{algorithm*}  

The main intuition behind Algorithm \ref{alg:naiveMWalgo} is that if views are mutually connected and we find sufficient number of high quality pairwise two-view redescriptions, there should exist those that can be successfully joined to form a complete multi-view redescription. 
Since there is no guided way in which incomplete redescriptions can be joined using this naive extension, Algorithm \ref{alg:naiveMWalgo} exhaustively tests all possible two-view redescriptions that can result in the increase of the number of views and finally in a completion of a redescription. 


\subsection{Additional evaluation measures}
\noindent In addition to the most-used redescription evaluation measures (presented in Section \ref{notation}), we use the following measures to evaluate redescriptions and redescription sets produced by the tested approaches. 

\par The attribute Jaccard index of two redescriptions, measuring the overall description redundancy of these redescription, is defined as: 
\begin{equation}
\label{measures:attJ}
attJ(R_1,R_2)= \frac{|attrs(R_1)\ \cap\ attrs(R_2)|}{|attrs(R_1)\ \cup\ attrs(R_2)|}
\end{equation}
\noindent Given a redescription set $\mathcal{R}$ containing $|\mathcal{R}|$ redescriptions, the average attribute Jaccard index of a redescription $R_i\in \mathcal{R}$ (measuring the average description redundancy of this redescription in the set) is defined as: 
\begin{equation}
\label{measures:avgAJ}
AAJ(R_i)=\sum_{R_j\in \mathcal{R},\ j\neq i} attJ(R_i,R_j)/(|\mathcal{R}|-1)
\end{equation} 
\noindent By analogy, the entity Jaccard index of two redescriptions is defined as: 
\begin{equation}
\label{measures:elemJ}
elemJ(R_1,R_2)= \frac{|supp(R_1) \cap supp(R_2)|}{|supp(R_1) \cup supp(R_2)|}
\end{equation}
 \noindent and the average entity Jaccard index as: 
\begin{equation} 
\label{measures:AvgEJ}
 AEJ(R_i)=\sum_{R_j\in \mathcal{R},\ j\neq i} elemJ(R_i,R_j)/(|\mathcal{R}|-1)
\end{equation} 
 \noindent These measures provide information about the redundancy of a redescription with respect to entities and attributes. 

Redescription complexity, given the number of attributes $k_c$ denoting complex queries, is computed as: 

\begin{equation}
   comp(R)=\left\{
\begin{array}{ll}
      |attr(R)|/k_c &, |attr(R)|< k_c\\
      1 &, k_c\leq |attr(R)| \\
\end{array} 
\right. 
\end{equation}

The aforementioned measures are naturally extended to scores used to evaluate sets of redescriptions (used in the experiments section of the manuscript). We use $AJ(\mathcal{R})$ to denote the average Jaccard index of all redescriptions contained in the redescription set $\mathcal{R}$. All measures are further transformed to have values in the $[0,1]$ range, so that $0$ denotes the best possible outcome and $1$ the worst possible outcome. Given $J_{sc}(R) = 1 - J(R)$:

\begin{equation}
J_{sc}(\mathcal{R}) = \sum_{i = 1}^{|\mathcal{R}|} (1-J(R_i))/|\mathcal{R}|
\end{equation}

\begin{equation}
A_{p_{sc}}(\mathcal{R}) = \sum_{i=1}^{|\mathcal{R}|}(log_{10}(p_{val}(R_i))/17+1.0)/|\mathcal{R}|
\end{equation}

Since $10^{-17}$ is the smallest $p_{value}$ we can compute exactly, $17$ is used as a normalization factor ($log_{10}(p_{val}(R_i))/17$ is used as a main part of $A_{p_{sc}}$).

\begin{equation}
AAJ_{sc}(\mathcal{R}) = \sum_{i = 1}^{|\mathcal{R}|} AAJ(R_i)/|\mathcal{R}| 
\end{equation}

\begin{equation}
AEJ_{sc}(\mathcal{R}) = \sum_{i = 1}^{|\mathcal{R}|} AEJ(R_i)/|\mathcal{R}| 
\end{equation}

\begin{equation}
comp_{sc}(\mathcal{R}) = \sum_{i = 1}^{|\mathcal{R}|} comp(R _i)/|\mathcal{R}| 
\end{equation}

\noindent The total score used to evaluate the redescription set is obtained as a weighted-sum function that integrates the aforementioned criteria with some predefined weights $w_i\in [0,1],\ \sum_{i=1}^{5} w_i = 1.0$. Intuitively, increasing the value of $w_i$ increases the importance of the measure $score_i$. 

\begin{multline}
total_{sc}(\mathcal{R}) = w_1\cdot J_{sc}(\mathcal{R}) + w_2\cdot A_{p_{sc}}(\mathcal{R}) + w_3\cdot AAJ_{sc}(\mathcal{R}) + \\ w_4\cdot AEJ_{sc}(\mathcal{R}) + w_5\cdot comp_{sc}(\mathcal{R}) 
\end{multline}

\noindent This combined score evaluates the quality of produced redescription set with respect to redescription accuracy, significance, entity and attribute redundancy and rule complexity. All measures used to compute $total_{sc}$ are decreasing, having value $0$ for the best possible redescription set (given a selected measure) and value $1$ for the worst possible redescription set. The total redescription set score is also decreasing achieving the same values $0$ ($1$) for the best (worst) possible redescription set given a combination of $5$ different measures. In barplots and tables throughout this manuscript we present a natural measure for redescription set accuracy: the average redescription set Jaccard index $J(\mathcal{R}) =  (\sum_{i = 1}^{|\mathcal{R}|} J(R_i))/|\mathcal{R}|$. This measure follows natural values of Jaccard index (which is used as a measure of redescription accuracy) achieving value $0$ for the least accurate set and value $1$ for the most accurate set of redescriptions. Explanations of the listed measures can be seen in \cite{mihelcic2017framework}.

All previously described redescription set measures are normalized by the size of the output redescription sets. Since it is possible that different approaches output redescription sets of different size, we have created scores that take this information into account. Approaches that do not succeed in producing the required number of redescriptions are penalized in these scores, because it is easier to create a small number of redescriptions satisfying the predefined criteria than doing the same with a larger number of redescriptions. We define an abstract redescription $R_{worst}$ such that $\forall i\ score_i(R_{worst}) = 1$. 

Given a user-defined number of desired (expected) output redescriptions $|\mathcal{R}_{out}|\in \mathbb{N}$  and any previously defined redescription  evaluation measure $score_i$. If $|\mathcal{R}_{out}|\geq|\mathcal{R}|$, the corresponding redescription set measure taking into account the number of actually produced redescriptions satisfying constraints is defined as:  

\begin{equation}
\begin{array}{l}
\underline{{score}_{i}}(\mathcal{R}) = (\sum_{i=1}^{|\mathcal{R}|} score_{i}(R_i) \vspace{1mm}\\
\hspace{2cm}+  \sum_{i=|\mathcal{R}|+1}^{|\mathcal{R}_{out}|} score_i(R_{worst}))/|\mathcal{R}_{out}|
\end{array}
\end{equation}

Redescription sets of a size $<|\mathcal{R}_{out}|$ are penalized since $|\mathcal{R}_{out}|-|\mathcal{R}|$ is added to the numerator of \underline{${score}_{i}$}. \underline{${score}_{i}$}($\mathcal{R}$) is a number in $[0,1]$. For sets such that $|\mathcal{R}|\geq|\mathcal{R}_{out}|$, $\underline{{score}_{i}}(\mathcal{R}) = score_i(\mathcal{R})$. The desired (expected) output redescription set size depends on the type of the analyses that is to be performed, level of knowledge required from the data, necessity of redescription validation by a domain expert etc. 

\subsection{Experimental setup}

The naive implementation of the multi-view redescription mining algorithm is realized using two general $2$-view redescription mining algorithms: the ReReMi \cite{GalbrunV:2012} and the CLUS-RM \cite{mihelcic2017framework}. The naive multi-view redescription mining algorithm using the ReReMi approach was run only once on each dataset, since the underlying algorithm creates the same set every time using some predefined set of input parameters. For the naive multi-view redescription mining algorithm using the CLUS-RM and all other presented approaches, we created $10$ different redescription sets, starting from different initial clusterings obtained by using randomizations with different seeds (see \cite{Mihelcic15LNAI}). The parameters used in each step of the evaluation process are listed in Table \ref{tab:parameters}, where we used identical parameters for the two-view CLUS-RM algorithm and identical constraints for the ReReMi algorithm. The framework for multi-view redescription mining and the two-view CLUS-RM algorithm were set to output maximally $200$ redescriptions. Since such restrictions do not exist for the ReReMi algorithm, all produced incomplete redescriptions were used in the naive approach using this method. Specific parameters for the ReReMi algorithm are provided in Section $S5$ of Supplementary document $1$. We used $\mathcal{S}.perc = 0.95$ in all runs of the naive algorithm. Unlike the proposed framework, it is not possible to explicitly control the size of the output redescription set produced by the naive approach (used in the $QP$ experiments). For the APP (see Section \ref{sub:app}), we create one redescription set by performing $10$ runs with the parameters specified in Table \ref{tab:parameters}. 

The presented barplots contain the average redescription set: Jaccard index (\underline{$J$}$(\mathcal{R})$), $p$-value score (\underline{$A_{p_{sc}}$}$(\mathcal{R})$), $\underline{{total}_{sc}}(\mathcal{R})$, execution time and memory consumption over $10$ runs or the exact value for the naive approach with the ReReMi algorithm.
The result tables of all experiments presented in this manuscript, containing the (average) performance measures achieved (over these $10$ runs) for each of the $12$ aforementioned redescription set measures and the corresponding standard deviations are available in the Supplementary document $1$. 

\begin{table*}[ht!]
\footnotesize
\caption{Algorithm parameters used to create redescription sets on the Country dataset (C), River water quality dataset (W) and Phenotype dataset (P). Different experiments are abbreviated as: QP - quality of produced sets, RFSM - random forest of supplementing models, RG - rule generation model, VRSP - view random subset projection, APP - application. For a thorough explanation of dataset $W_p$ see Section \ref{sub:app}.}
\label{tab:parameters}
\centering
\begin{tabular}{ |c|c|c|c|c|c|c|c|c|c|c| }
\hline
$Exp.$ & $\mathcal{D}$ & $J_{minA}$ &$J_{min}$ & $p_{max}$ & supp. & $|r_i|$ & Iter. & $\mathcal{L}$ & $|\mathcal{T}|$ & $|\mathcal{R}_{out}|$ \\
\hline

& $C$ & $0.5$ & $0.6$ & $0.01$ & $[5,100]$ & $8$ & $5$ & all & $1$ & $200$ \\
QP/VRSP & $W$ & $0.3$ & $0.5$ & $0.01$ & $[5,800]$  & $8$ & $5$ & all & $1$ & $200$\\
& $P$ & $0.01$& $0.1$ & $0.01$ & $[5,70]$  & $8$ & $15$ & all & $1$ & $200$\\ \hline
& $C$ & $0.5$ & $0.6$ & $0.01$ & $[5,100]$ & $4$ & $2$ & all & $1+\{20,50\}$ & $200$\\
RFSM & $W$ & $0.3$ & $0.6$ & $0.01$ & $[5,800]$  & $8$ & $4$ & all & $1+\{20,50\}$ & $200$\\
& $P$ & $0.3$ & $0.5$ & $0.01$ & $[5,70]$  & $4$ & $15$ & all & $1+\{20,50\}$ & $200$\\ \hline
& $C$ & $0.5$ & $0.6$ & $0.01$ & $[5,100]$ & $5$ & $2$ & all & $1,2,4,6$ & $200$\\
RG & $W$ & $0.3$& $0.6$ & $0.01$ & $[5,800]$  & $8$ & $4$ & all & $1,2,4,6$ & $200$\\
& $P$ & $0.2$ & $0.4$ & $0.01$ & $[5,70]$  & $6$ & $15$ & all & $1,2,4,6$ & $200$\\ \hline
APP & $W_p$ & $0.01$ & $0.2$ & $0.01$ & $[5,150]$  & $8$ & $10$ & all & $1$ & $200$\\

\hline
\end{tabular}
\end{table*}

\noindent $J_{minA}$ denotes the minimal Jaccard index required to use redescriptions in the conjunctive refinement procedure (see \cite{mihelcic2017framework}), $|r_i|$ is the maximal rule length obtainable by transforming PCTs to rules and the average rule length obtainable by transforming forest of tree-based models to rules, $\mathcal{L}$ ($op$) denotes the query language (conjunction, disjunction, negation, all operators), $\mathcal{T}$ the number of trees used in the experiment and $Iter.$ the number of iterations used in the GCLUS-RM algorithm. Notation $1+\{20,50\}$ denotes two settings, $1$ PCT with a supplementing forest model containing either $20$ or $50$ trees. To demonstrate the difference between using a single PCT model and the PCT model supplemented by a Random Forest of models, we increased the strictness of accuracy constraints in these experiments. To reduce the overall execution time of these experiments and achieve faster redescription generation when using supplementing models, we reduced the number of iterations and tree depths as compared to the experiments using a single PCT model. In all experiments we used $k_c = 20$, the expected redescription set size $|\mathcal{R}_{out}| = 200$, and the $1\times 5$ matrix $\mathcal{W}$ having all entries equal to $0.2$. The expected redescription set size is set to $200$, since this is large enough for contained redescriptions to capture quality share of knowledge contained in the data, it is large enough to allow performing statistical analyses, but  is small enough to be examined by the domain experts in a reasonable time. 


\subsection{Evaluating quality of produced redescription sets}

We present the results of evaluating our framework for multi-view redescription mining using one PCT as rule-generating model and compare these results with the naive algorithms for multi-view redescription mining. Statistical significance of the difference of mean value of different redescription quality measures, between the proposed framework and the naive method using CLUS-RM two-view redescription mining algorithm, was computed using one-sided Mann-Whitney U test. The test assesses if the mean value achieved by the proposed multi-view redescription mining framework is significantly larger than achieved by the naive extension using CLUS-RM for the \underline{$J$}$(\mathcal{R})$ and if it is significantly lower for other measures. Comparative barplots presented in the manuscript contain evaluation results of the framework using $(1000, 4000)$ memory configuration. The second experiment compares the performance of the proposed framework using different values of memory parameters (working and maximal allowed memory size). Detailed table can be seen in Supplementary document $1$ (Table $S1$). 

\begin{figure*}[ht!]
 \centering
    \subfloat[ \underline{${J}$}($\mathcal{R}$), \underline{${A_{p_{sc}}}$}($\mathcal{R}$), \underline{${total}_{sc}$}($\mathcal{R}$).]{{\includegraphics[width=8cm]{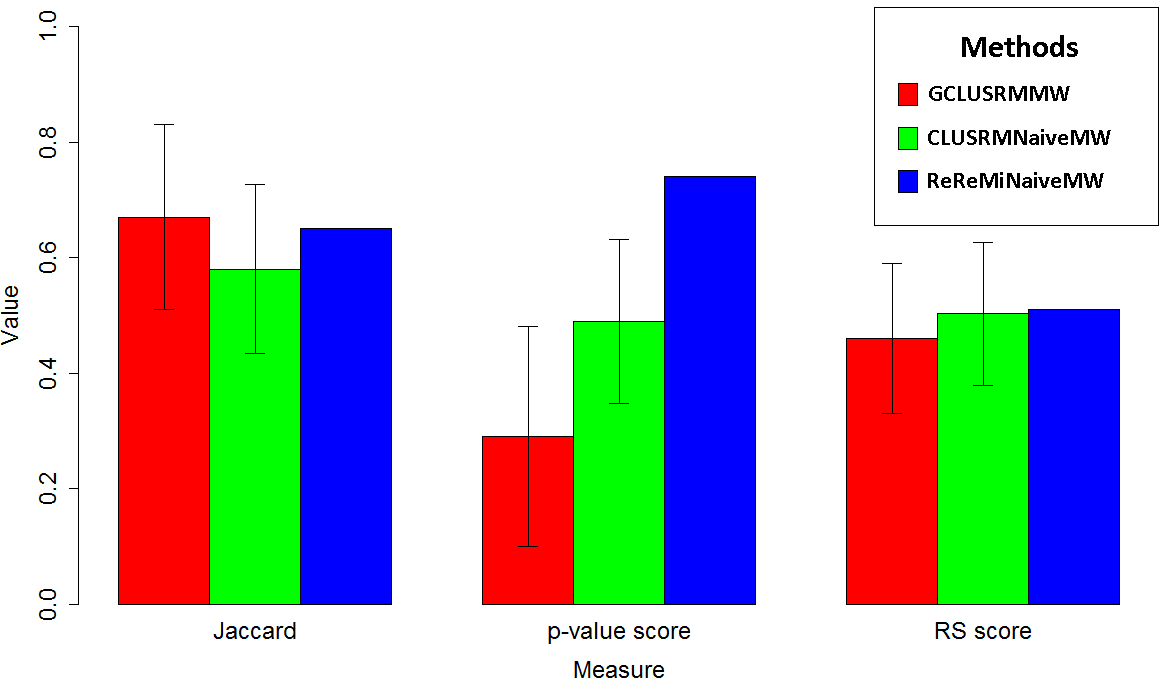} }}%
    \qquad
    \subfloat[Average execution time in seconds.]{{\includegraphics[width=8cm]{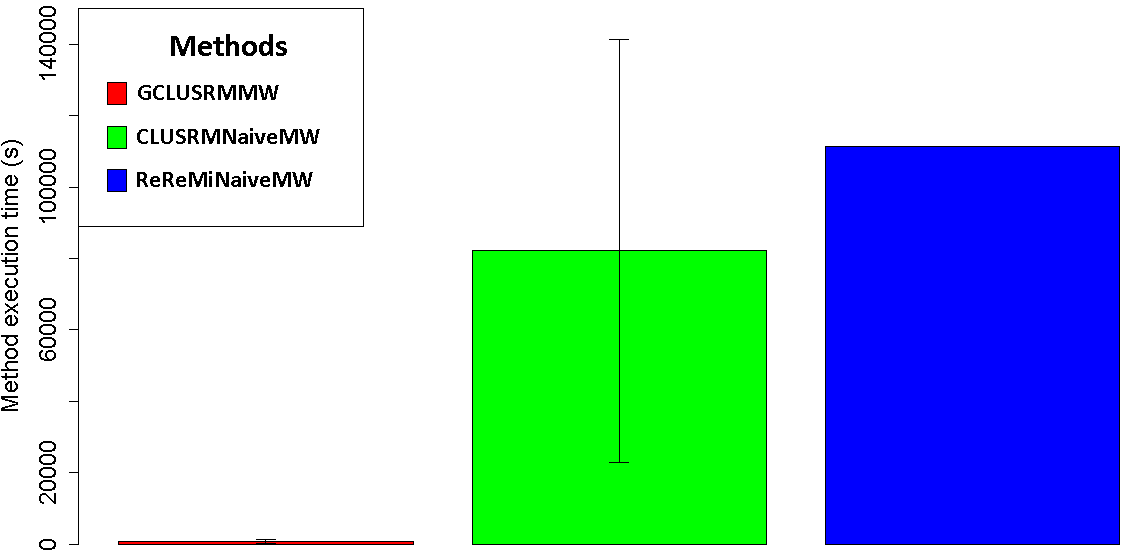} }}%
     \qquad
    \subfloat[Maximum measured RAM memory consumption in MB.]{{\includegraphics[width=8cm]{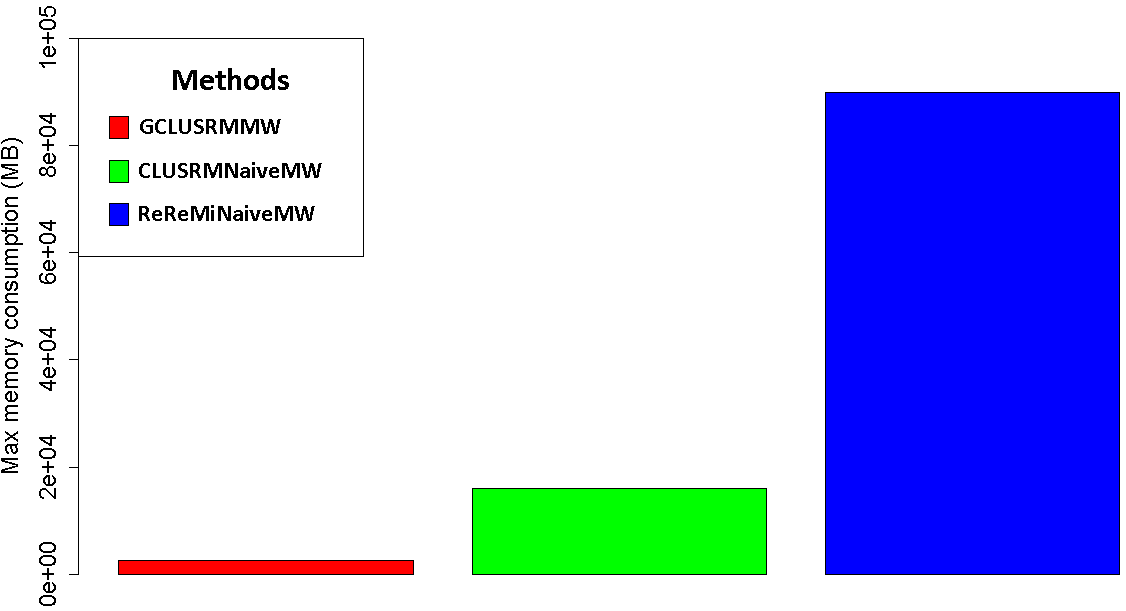} }}%
    \caption{Comparison results on the Country dataset.}%
    \label{fig:histsCountry}%
\end{figure*}

\noindent Comparative evaluation results presented in Fig. \ref{fig:histsCountry} show that the proposed multi-view redescription mining framework outperforms the naive algorithms with respect to redescription accuracy ($p = 5\cdot 10^{-4}$), significance ($p = 5\cdot 10^{-4}$), overall redescription set score ($p = 0.0019$), execution time ($p = 5.4\cdot 10^{-6}$) and maximal memory. Country dataset is the prime example why naive extension is not suitable for general use in multi-view redescription mining. The fact that it is possible to find large number of accurate $2$-view redescriptions on this dataset necessitates more elaborate techniques for pattern pruning and selection. 

\begin{figure*}[ht!]
 \centering
    \subfloat[ \underline{${J}$}($\mathcal{R}$), \underline{${A_{p_{sc}}}$}($\mathcal{R}$), \underline{${total}_{sc}$}($\mathcal{R}$).]{{\includegraphics[width=8cm]{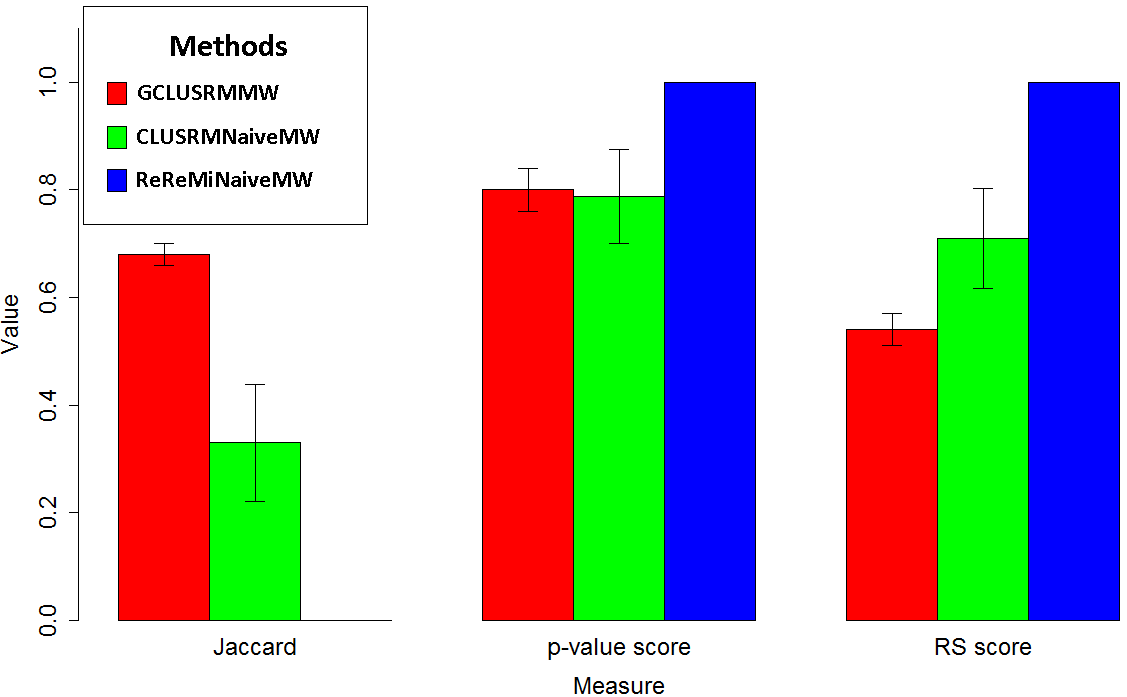} }}%
    \qquad
    \subfloat[Average execution time in seconds.]{{\includegraphics[width=8cm]{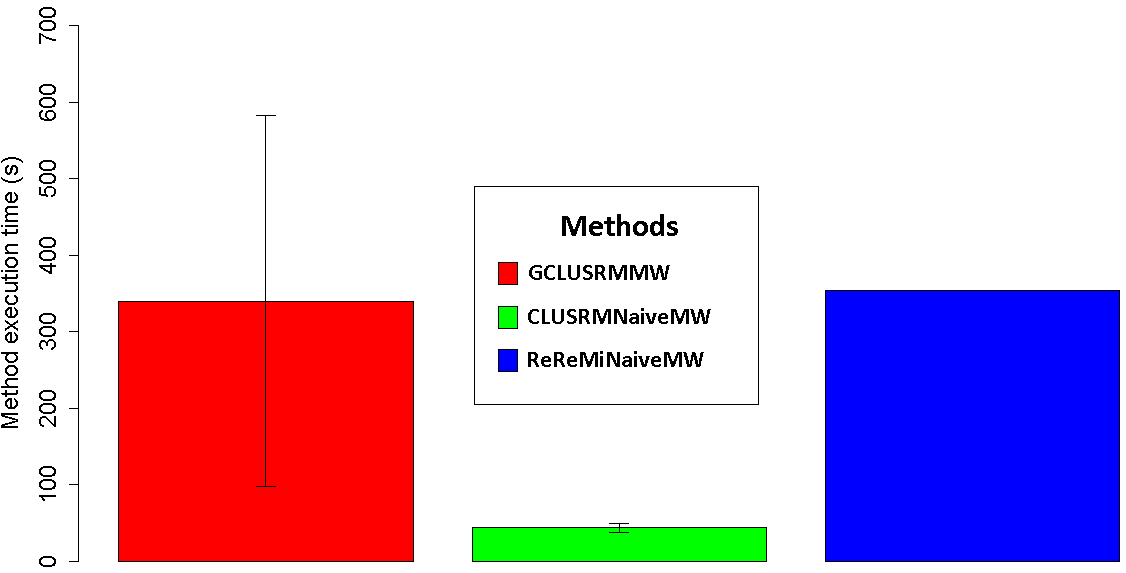} }}%
     \qquad
    \subfloat[Maximum measured RAM memory consumption in MB.]{{\includegraphics[width=8cm]{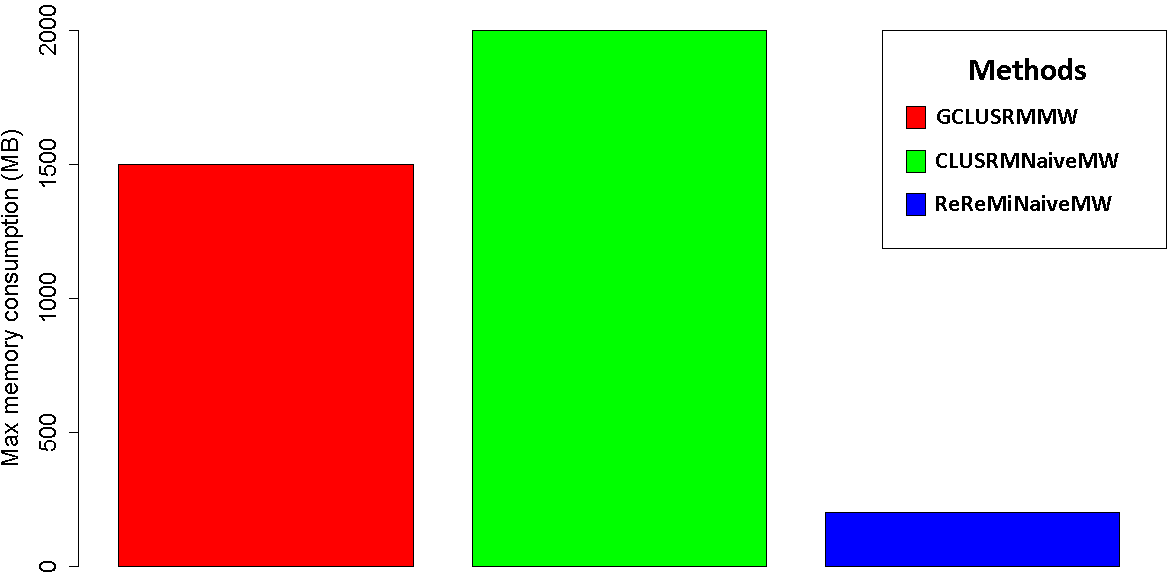} }}%
    \caption{Comparison results on the Slovenian Water dataset}%
    \label{fig:histsSloWater}%
\end{figure*}

\noindent If the underlying data does not allow creating many accurate redescriptions (as is the case for the Slovenian Water dataset and in part with Phenotype dataset) the execution time of the naive algorithm can be even smaller than the execution time of the proposed framework due to smaller number of applications of the $2$-view techniques - ${n \choose 2}$ compared to $>n\cdot {n \choose 2}$ (see supplementary document $1$, Section S$3$ for more detailed explanation). Results in Fig. \ref{fig:histsSloWater} show that the proposed multi-view framework outperforms naive implementation with respect to redescription accuracy ($p=5.4\cdot 10^{-6}$) and overall redescription set score ($p = 3.79\cdot 10^{-5}$). The difference in mean of redescription statistical significance score, between the proposed framework and the naive approach using the CLUS-RM algorithm, is not significant. Naive implementation with CLUS-RM also uses larger maximum amount of memory. Naive approach using ReReMi algorithm did not manage to produce any satisfactory multi-view redescriptions on this dataset which explains overall low memory consumption.

\begin{figure*}[ht!]
 \centering
    \subfloat[ \underline{${J}$}($\mathcal{R}$), \underline{${A_{p_{sc}}}$}($\mathcal{R}$), \underline{${total}_{sc}$}($\mathcal{R}$).]{{\includegraphics[width=8cm]{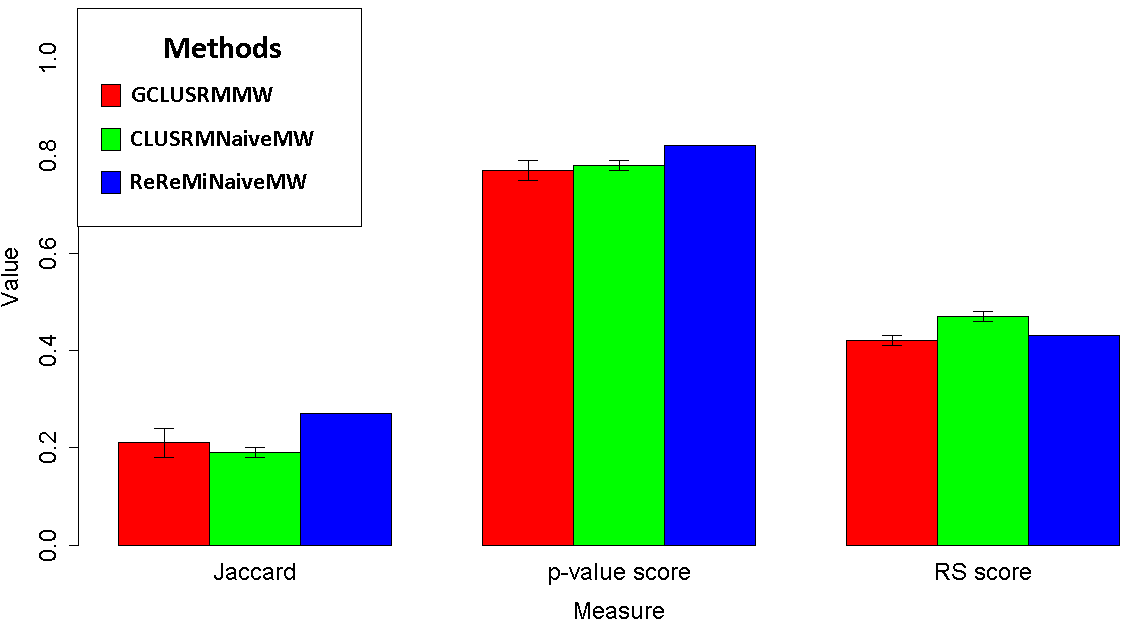} }}%
    \qquad
    \subfloat[Average execution time in seconds.]{{\includegraphics[width=8cm]{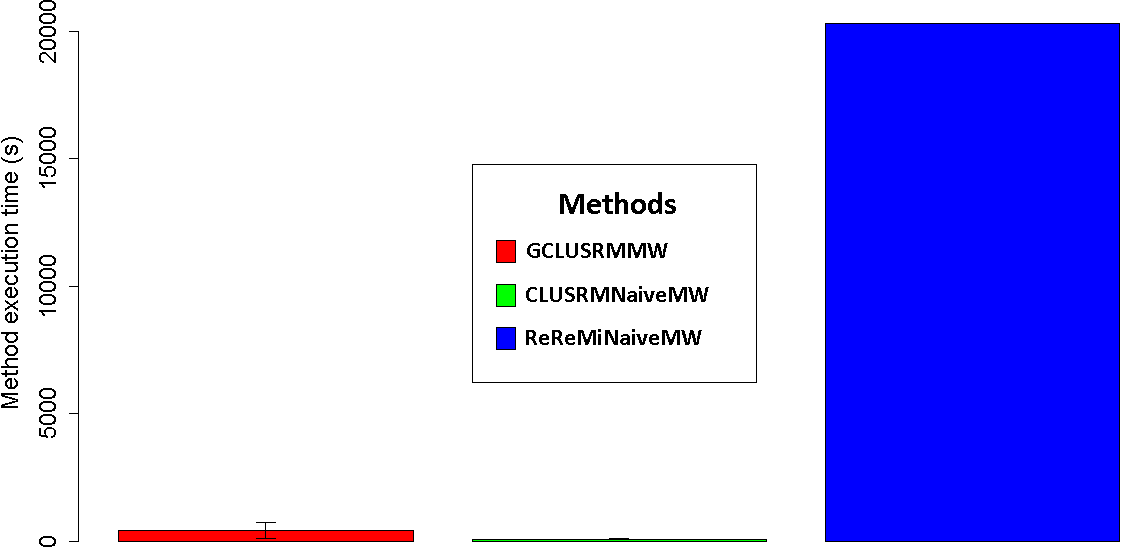} }}%
     \qquad
    \subfloat[Maximum measured RAM memory consumption in MB.]{{\includegraphics[width=8cm]{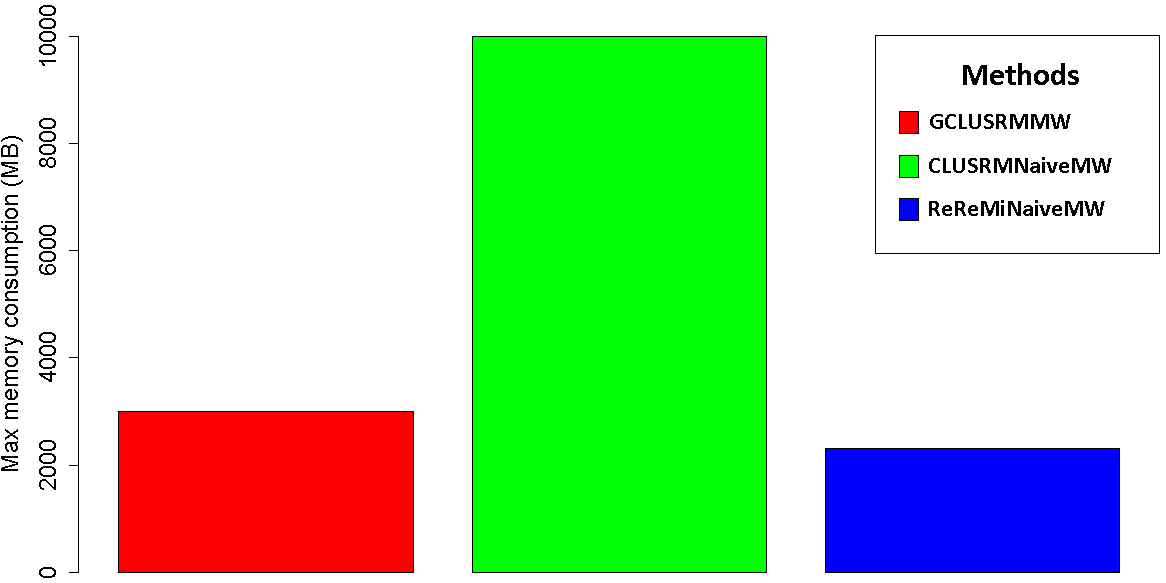} }}%
    \caption{Comparison results on the Phenotype dataset}%
    \label{fig:histsPheno}%
\end{figure*}

\noindent Comparison results on the Phenotype dataset, presented in Fig. \ref{fig:histsPheno}, show that the naive approach using ReReMi $2$-view algorithm produces the most accurate redescriptions, whereas the proposed framework produces at average the most significant redescriptions (difference in mean value compared to the naive approach with CLUS-RM is not significant) and redescription sets with the best overall redescription set score ($p = 5.4\cdot 10^{-6}$). It has significantly smaller execution time than the naive implementation using the ReReMi algorithm and significantly smaller maximal memory consumption than the naive approach using the CLUS-RM algorithm. Information about entity and attribute coverage of produced redescription sets for all presented methods is available in Supplementary document $1$.

Overall, the proposed framework offers a well-balanced trade-off between accuracy, memory consumption and execution time. It mostly outperforms the naive approach with respect to accuracy, execution time and memory consumption. Most importantly, unlike the naive approach that depends on the properties of the underlying data, the proposed framework is a generally applicable methodology.

\begin{figure*}[ht!]
\centerline{\includegraphics[width=1.0\textwidth]{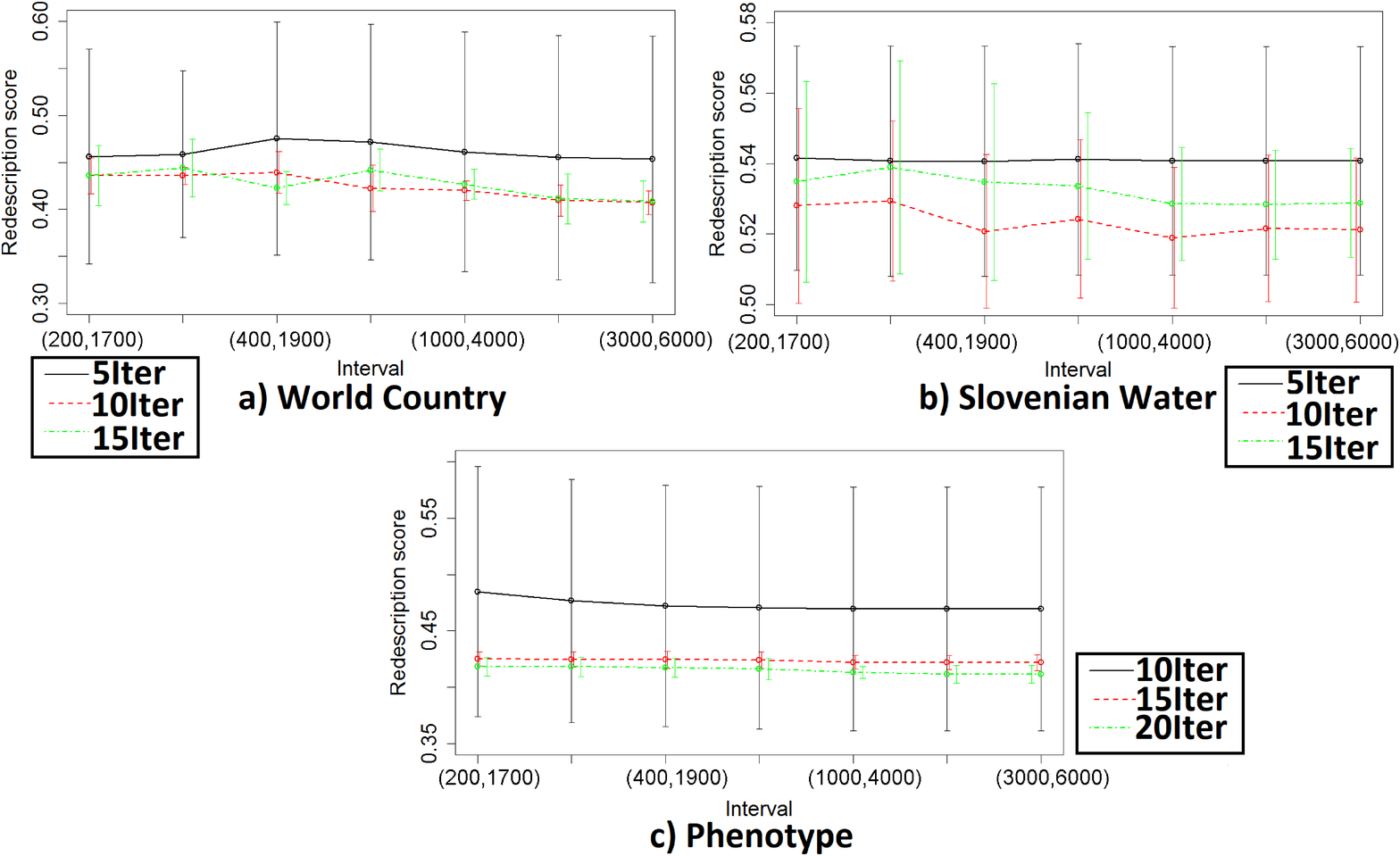}}
\caption{Overall score $\underline{total_{sc}}(\mathcal{R})$ obtained for the redescription set using a single rule-generating PCT model with different memory and iteration parameters.}
\label{fig:STRS}
\end{figure*}

The results presented in Table $S1$ of Supplementary document $1$ and Fig. \ref{fig:STRS} show that increasing the amount of memory allows obtaining redescription sets of higher quality. The difference in quality between the redescription set produced with the smallest memory setting ($200,\ 1700$) and the largest memory setting ($3000,\ 6000$) across $10$ runs is not significant with $5$ iterations for the Country ($p = 0.423$) and the Water ($p = 0.278 $) dataset but it is significant for the Phenotype dataset ($0.032$) with the significance level $0.05$. Increasing the amount of iterations used to create redescriptions mostly improves this result.  This difference in accuracy is statistically significant on the Country dataset with $10$ ($p=9.77\cdot 10^{-4}$) and $15$ ($p=0.0244$) iterations and on the Phenotype dataset with $10$ ($p=0.049$), $15$ ($p=0.032$) and $20$ ($p=0.003$) iterations. Although the difference is not significant on the Slovenian Water dataset, increasing the number of iterations increases the difference in accuracy between the first and the last memory configuration. Statistical significance was measured using Wilcoxon signed-rank test. The final redescription set score is very stable on the Slovenian Water and Phenotype datasets, but its standard deviation is significantly higher on the World Country dataset. The main reason for this is the inability of the framework to produce $200$ complete redescriptions for each of the $10$ different runs. Increasing the number of iterations ($Iter.$) resolves this problem. 

The execution time analyses of the framework for multi-view redescription mining with various number of views is available in Section $S3$ of Supplementary document $1$.

\subsection{Using Random Forest of supplementing models}

This section evaluates the use of a Random Forest of supplementing models inside a framework for multi-view redescription mining.
Detailed analyses of the performed experiments can be seen in Section $1.2$ of Supplementary document $1$.

\begin{figure*}[ht!]
\centerline{\includegraphics[width=1.0\textwidth]{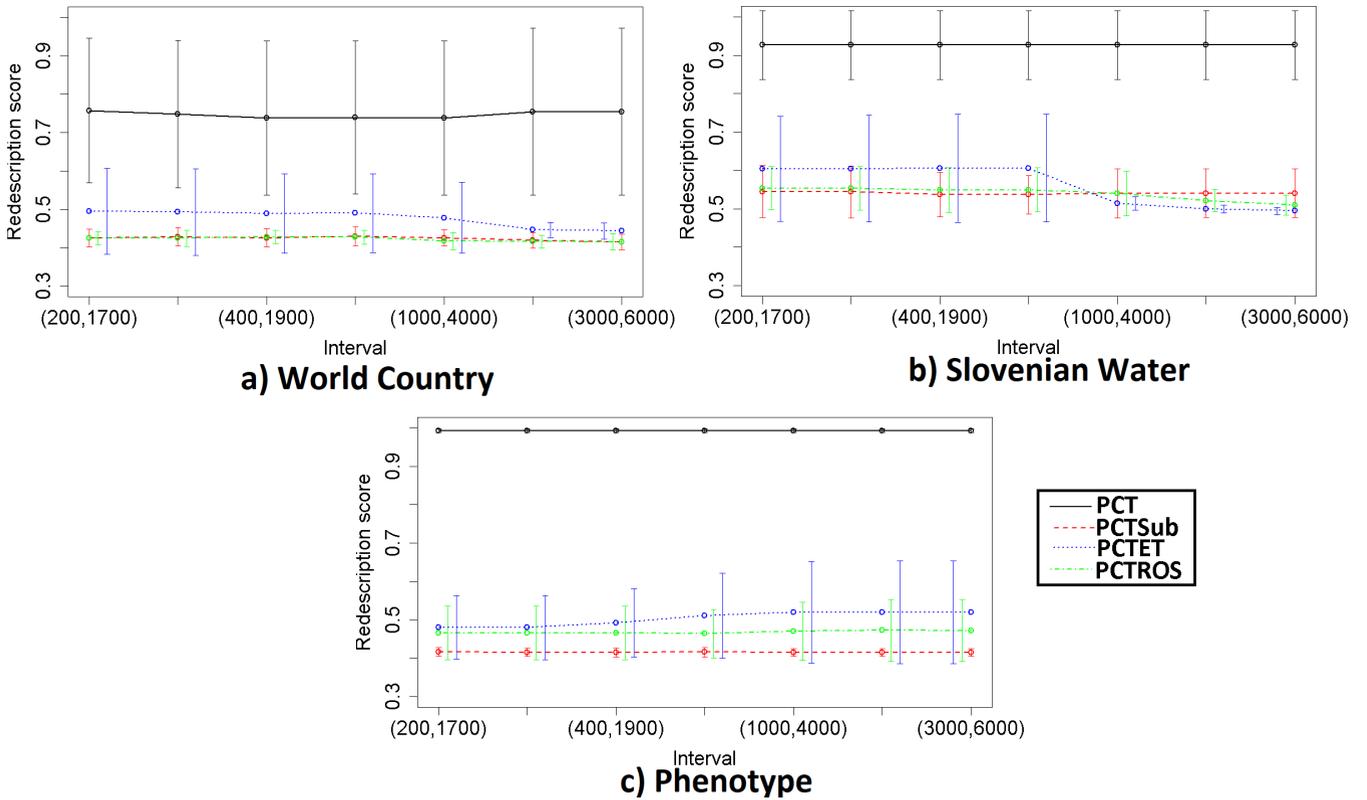}}\caption{Overall redescription set score $\underline{total_{sc}}(\mathcal{R})$ obtained using a single rule-generating PCT model and a supplementing model containing $50$ trees with different memory parameters.}
\label{fig:SM}
\end{figure*}

It is visible from Tables $S2-S7$ of Supplementary document $1$, as well as Fig. \ref{fig:SM}, that using supplementing models significantly increases the performance of the proposed framework (both with respect to accuracy and stability). The corresponding $p$-values of the difference in mean value of the average redescription set score achieved using a supplementing model compared to using only a single rule-generating PCT at each memory setting, according to the one-sided Wilcoxon signed-rank test, are: $p_{(PCT_{Sub},PCT)}= 7.8\cdot 10^{-3} $, $p_{(PCT_{ET}, PCT)}= 7.8\cdot 10^{-3} $, $p_{(PCT_{ROS}, PCT)}= 7.8\cdot 10^{-3}$ on the World Country dataset, $p_{(PCT_{Sub},PCT)}= 0.011 $, $p_{(PCT_{ET}, PCT)}= 7.8\cdot 10^{-3} $, $p_{(PCT_{ROS}, PCT)}= 7.8\cdot 10^{-3}$ on the Slovenian Water dataset and $p_{(PCT_{Sub},PCT)}= 7.8\cdot 10^{-3} $, $p_{(PCT_{ET}, PCT)}= 0.011 $, $p_{(PCT_{ROS}, PCT)}= 7.8\cdot 10^{-3}$ on the Phenotype dataset. 
Fig. \ref{fig:SM} shows that increasing the memory parameters increases performance of the framework with supplementing model on the World Country and the Slovenian Water dataset, while the performance slightly degrades when the Extra multi-target PCTs are used as a supplementing model, with the increased memory parameters, on the Phenotype dataset. This occurs due to the model inability to produce $200$ redescriptions at each run.

\subsection{Extra trees as a main rule generation model}

In this section, we consider the overall performance of a framework for multi-view redescription mining when using one or more Extra multi-target PCTs as a main rule-generating model as compared to using one Predictive Clustering tree as a main rule-generating model.

The experiments presented in Fig. \ref{fig:histsGen} and Table $S8$ of Supplementary document $1$ demonstrate that using a few Extra multi-target PCTs as the main rule-generating model in the proposed multi-view redescription mining framework can significantly outperform using one main rule-generating PCT model. Using the Extra multi-target PCTs as a main rule-generating model increases the overall performance, accuracy, diversity and the number of produced redescriptions (given a set of predefined constraints defined in Section \ref{evaluation}). This is visible from the $\underline{underline}$ scores and corresponding standard deviations presented in Fig. \ref{fig:histsGen}. Given the fact that learning multiple Extra multi-target PCTs in parallel can be achieved easily on any modern PC, the overall gain can be substantial. More detailed analyses can be seen in Section $S1.3$ of Supplementary document $1$.

\begin{figure*}[ht!]
 \centering
    \subfloat[\underline{${J}$}($\mathcal{R}$), \underline{${A_{p_{sc}}}$}($\mathcal{R}$), \underline{${total}_{sc}$}($\mathcal{R}$) on the Country dataset.]{{\includegraphics[width=8cm]{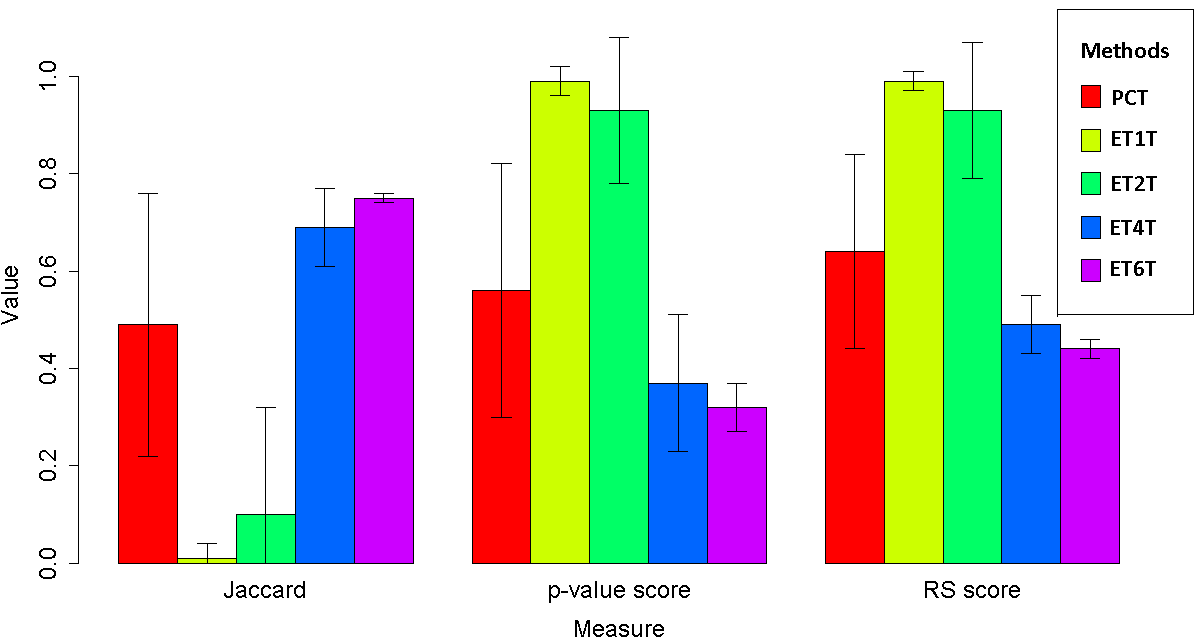} }}%
    \qquad
    \subfloat[\underline{${J}$}($\mathcal{R}$), \underline{${A_{p_{sc}}}$}($\mathcal{R}$), \underline{${total}_{sc}$}($\mathcal{R}$) on the Slovenian Water dataset.]{{\includegraphics[width=8cm]{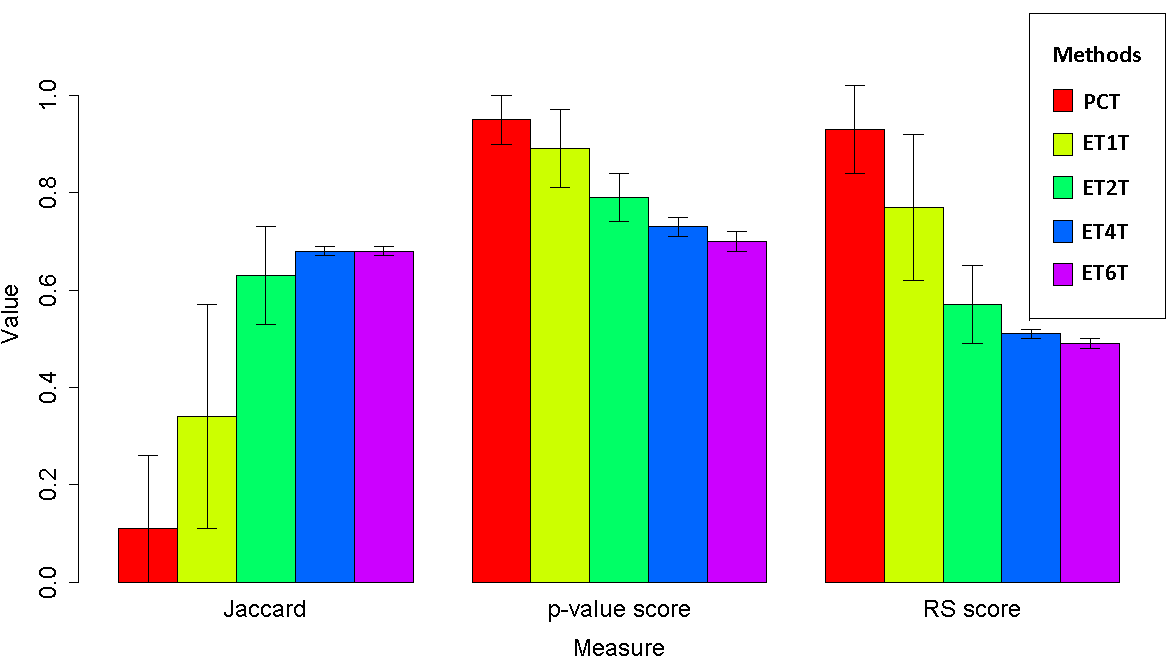} }}%
     \qquad
    \subfloat[ \underline{${J}$}($\mathcal{R}$), \underline{${A_{p_{sc}}}$}($\mathcal{R}$), \underline{${total}_{sc}$}($\mathcal{R}$) on the Phenotype dataset.]{{\includegraphics[width=8cm]{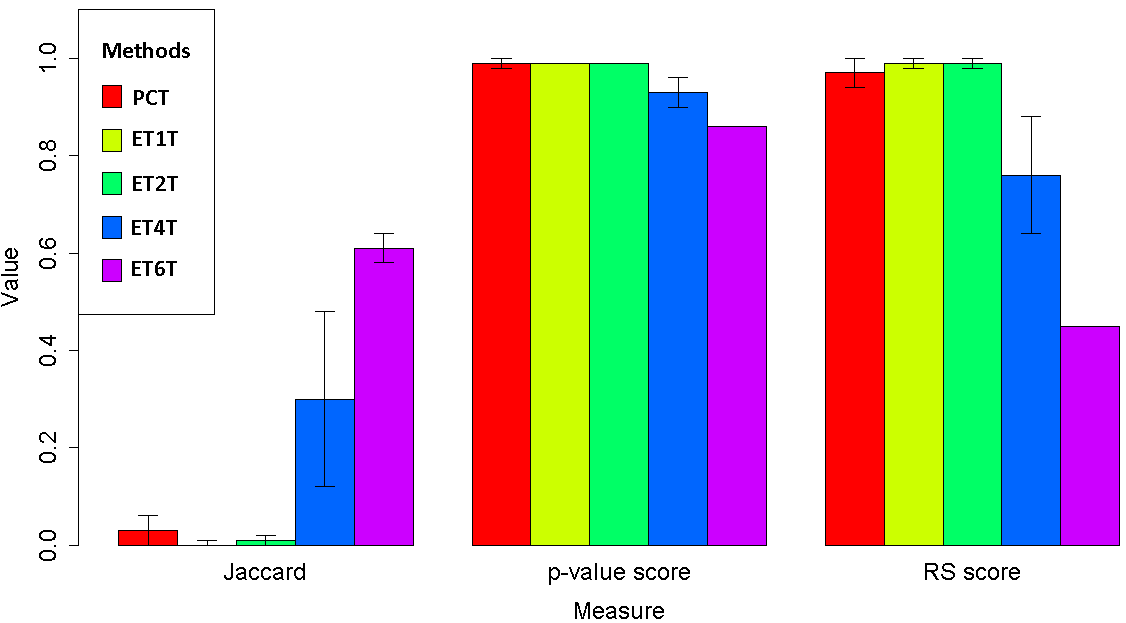} }}%
    \caption{The framework's performance using one main PCT rule-generating model compared to using $1,\ 2,\ 4$ or $6$ Extra multi-target PCTs (ET) as a main rule-generating model. A working set size of at most $3000$ and a maximal memory size of $6000$ are used in all experiments.}%
    \label{fig:histsGen}%
\end{figure*}

\subsection{View random subset projections}
The proposed framework applies the CLUS-RM algorithm to each pair of available views and then completes the obtained incomplete redescriptions using rules produced on the remaining views. Although less computationally complex than the naive generalization of redescription mining algorithms, this approach requires performing $n \choose 2$ CLUS-RM applications and the same number of redescription completions (where $n$ denotes the number of available views). We test how much is lost by using and completing only a fixed size subset of pairs of initial views - performing random view subset projection. In the experiments performed to obtain results presented in Fig. \ref{fig:histsProj} and Table $S9$ of Supplementary document $1$, we used $2$ pairs of initial views to create redescriptions. Thus, performing such a random view subset projection executes $3$ times faster on the World Country dataset and $33\%$ faster on the Slovenian Water and the Phenotype dataset as compared to a regular run of the multi-view redescription mining framework.

The results presented in Fig. \ref{fig:histsProj} and Table $S9$ of Supplementary document $1$ show that, expectedly the full run of the multi-view redescription mining framework outperforms the random view subset projection runs. However, the difference in average redescription set score (after performing $10$ runs) is $0.1$ ($10\%$ of redescription set score range, or full run obtains a set that has $\sim 18\%$ better score than obtained by projection) on the World Country dataset, $0.06$ ($6\%$ of redescription set score range, or $\sim 7\%$ better) on the Slovenian Water dataset and $0.01$ ($1\%$ of redescription set score range, or $\sim 3\%$ better) on the Phenotype dataset. Higher deviation between runs when using projection (approx. $2\times$ on Country, $5\times$ on Water and almost identical on Phenotype dataset) must be taken into account. However, if multiple runs are used, as was done in this experiment (to reduce the effects of  deviation), random view subspace projection may  be used as a technique to alleviate the curse of dimensionality in multi-view redescription mining.

\begin{figure*}[ht!]
 \centering
    \subfloat[ \underline{${J}$}($\mathcal{R}$), \underline{${A_{p_{sc}}}$}($\mathcal{R}$), \underline{${total}_{sc}$}($\mathcal{R}$) on the Country dataset.]{{\includegraphics[width=8cm]{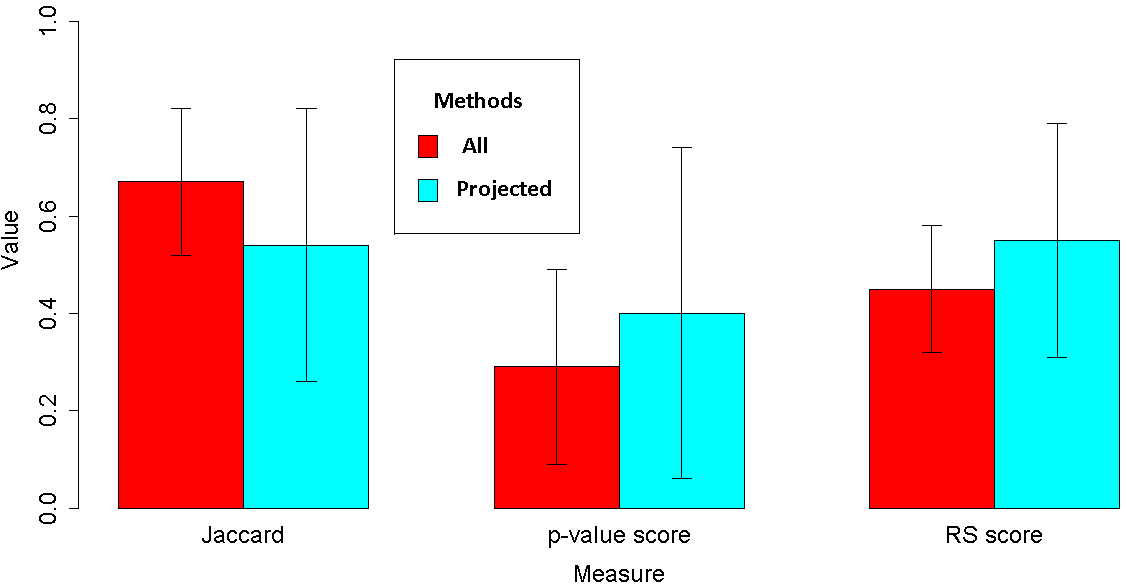} }}%
    \qquad
    \subfloat[\underline{${J}$}($\mathcal{R}$), \underline{${A_{p_{sc}}}$}($\mathcal{R}$), \underline{${total}_{sc}$}($\mathcal{R}$) on the Slovenian Water dataset.]{{\includegraphics[width=8cm]{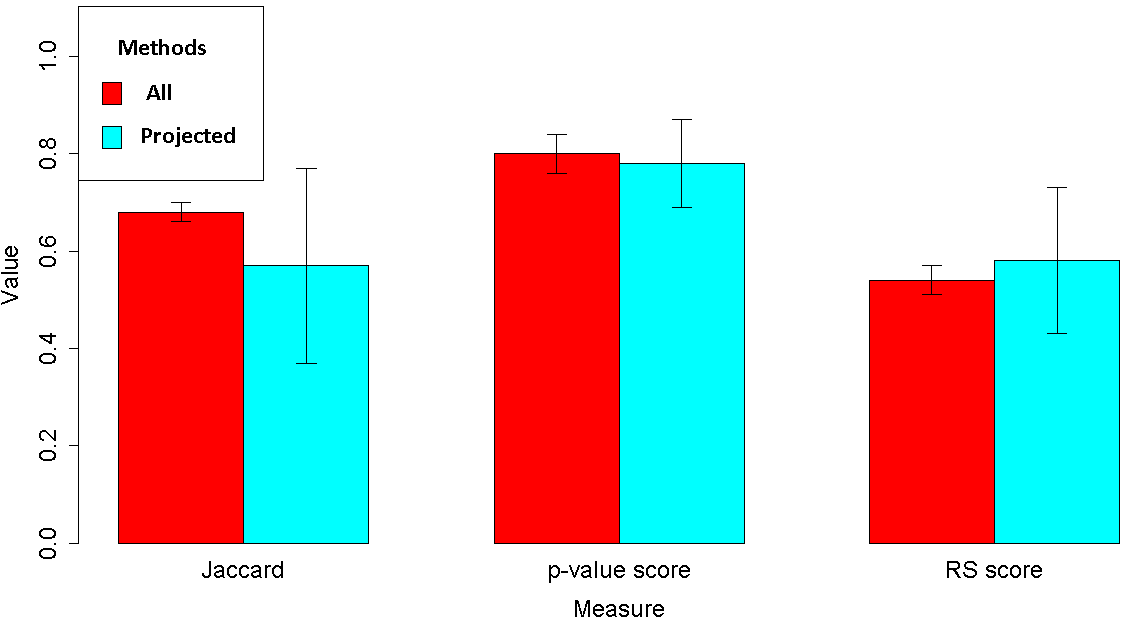} }}%
     \qquad
    \subfloat[ \underline{${J}$}($\mathcal{R}$), \underline{${A_{p_{sc}}}$}($\mathcal{R}$), \underline{${total}_{sc}$}($\mathcal{R}$) on the Phenotype dataset.]{{\includegraphics[width=8cm]{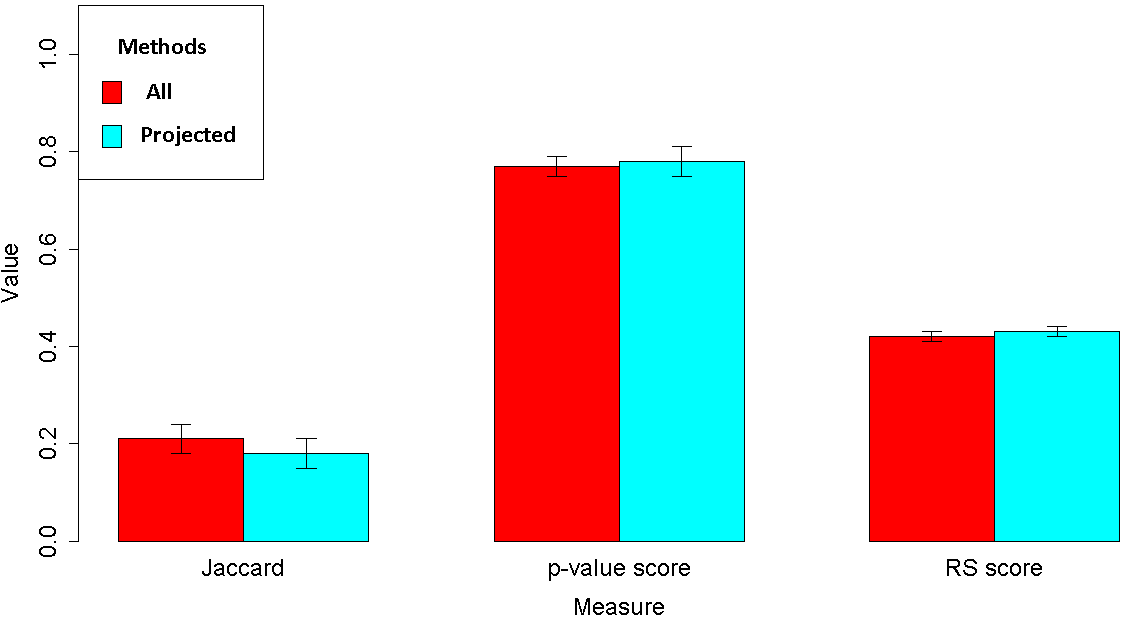} }}%
    \caption{The evaluation of redescription sets created using two pairs of initial views on the World Country (C), Water Quality (W) and Phenotype (P) datasets.}%
    \label{fig:histsProj}%
\end{figure*}

\subsection{Application -  understanding of machine learning models}
\label{sub:app}
In this subsection, we show the benefits of using multi-view redescription mining in machine learning. Namely, it can increase the understanding of any machine learning model and interrelate predictions made by a set of different machine learning models. We also show how to gain new knowledge by incorporating information obtained from machine learning models into multi-view redescription mining setting.

We use the Phenotype dataset, described in Section \ref{data}, to demonstrate the use of feature ranking, obtained by the Random Forest algorithm, in the multi-view redescription mining setting. This allows relating feature importance's for predicting bacterial phenotypes of features obtained on different sets of attributes such as:  metagenomic co-occurrences, proteome composition and genomic signatures of translation efficiency in gene families. Using this type of analyses reveals properties about the underlying problem but also about the model used to create the ranking (since observed redescriptions summarise the model output). In our Phenotype use-case dataset, redescription mining allows detecting subsets of phenotypes and the corresponding subsets of features that are predictive (to some degree) for all phenotypes in a given subset. This information is not easily deducible from the phenotype-specific long lists of feature importance scores. The approach allows relating feature importance scores obtained using multiple different models or feature ranking approaches. It also allows relating model output (such as feature ranking) to the original sets of features, which provides additional information to the domain experts (allows describing discovered subsets of phenotypes that share common informative features using original attribute value range). The insights provided by the approach can be used to make corrections, parameter tuning or model selection. If highly accurate model is used, obtained feature scores can be used as a filter to constrain search for redescriptions only on the selected subset of predictive attributes (this can be achieved by applying stricter threshold prior to application of the redescription mining algorithm) . This significantly reduces execution time and eliminates many potentially uninteresting patterns (reduces the possibility of finding subsets of phenotypes that share predictive features of low or medium importance). All this makes the overall analyses more efficient.

To demonstrate the use of multi-view redescription mining to relate and understand predictions made by  multiple machine learning models and incorporate these predictions into the redescription mining setting, we create a new dataset derived from the Slovenian Water dataset.  We use the physical and chemical measurements of water quality as attributes and predict the occurrence of $7$ different plant and $7$ different animal species in waters from different locations in Slovenia. First, we randomly shuffle the data and make $70\%-30\%$ split. We train a Random Forest of $600$ Predictive Clustering trees with and without random output selection and a Random Forest of $600$ Extra multi-target PCTs on a train set to predict the occurrence of aforementioned species (these are used due to their multi-label classification and multi-target regression abilities but in principle any model can be used).  Our multi-view redescription mining dataset is comprised of the $30\%$ test split having $4$ views. The first view is the corresponding part of the Slovenian Water dataset containing physical and chemical measurements for corresponding locations contained in the test split. The remaining three views are predictions obtained by the aforementioned models (integer $\{0,1,3,5\}$, where $0$ represents no occurrence and $5$ represents the abundant occurrence of the organism). 

Using such datasets brings many advantages and benefits for machine learning, explainable data science and redescription mining. Since it allows relating predictions made by the multiple approaches it effectively allows combining approaches in a non-linear way, also allowing to understand the similarities and the differences of obtained predictions for various subgroups of data. Since it allows describing the obtained subgroup with original attributes, it provides means for analyses and verification by the domain experts. Further, it allows detecting subgroups on which models make mistakes or on which models disagree providing the interpretable justification for model tuning or selection. Finally, the obtained redescriptions can be used to locate subsets of unseen examples on which it is highly expected that some predefined property holds or they can be used as more complex yet interpretable local predictors. There are also benefits of adding views obtained by the machine learning models into the multi-view redescription mining (when target labels are available). It is often very hard to segment the data in the completely unsupervised manner (only using original attribute values). Adding one or more views containing predictions of machine learning models allows focusing redescription creation to these redescriptions describing one or more target classes of interest. Such a procedure can also be applied when the target labels are available, however, using machine learning models is a more general approach allowing for focused redescription mining on both the annotated and the unannotated part of data. 
 
\begin{center}
\begin{footnotesize}
\LTcapwidth=\textwidth
\begin{table*} 
\caption{Example redescriptions illustrating the use of multi-view redescription mining to interpret machine learning models and incorporating information obtained by these models to gain new insights in redescription mining. These redescriptions have been obtained on the Phenotype dataset (pheno) and the modified Slovenian Water dataset (sw).}
\label{tab:redgEx}
\begin{tabular}{l l}\\
$R_{pheno_1}:$ & $(q_{1_{pheno_1}},q_{2_{pheno_1}}, q_{3_{pheno_1}})$\\
$q_{1_{pheno_1}}:$ & $5\cdot 10^{-5} \leq \text{\texttt{taxID}}_{332410}\leq 2.3\cdot 10^{-4}$ \\
$q_{2_{pheno_1}}:$ & $2.6\cdot 10^{-3} \leq \text{\texttt{CE}}\leq 3.5\cdot 10^{-3}\ \wedge\  1.4\cdot 10^{-3} \leq \text{\texttt{VG}} \leq 2.2\cdot 10^{-3}$ \\
$q_{3_{pheno_1}}:$ & $1.6\cdot 10^{-4} \leq \text{\texttt{COG}}_{583}\leq 9.1\cdot 10^{-4}\ \wedge\  9.3\cdot 10^{-4} \leq \text{\texttt{COG}}_{209} \leq 1.1\cdot 10^{-3}$ \\ 
quality: & $J(R_{pheno_1}) = 1.0,\ |supp(R_{pheno_1})| = 5,\ p(R_{pheno_1}) = 1.2\cdot 10^{-14}$\\
& \\

$R_{pheno_2}:$ & $(q_{1_{pheno_2}},q_{2_{pheno_2}}, q_{3_{pheno_2}})$ \\
$q_{1_{pheno_2}}:$ & $\neg (1\cdot 10^{-6} \leq \text{\texttt{taxID}}_{358220}\leq 1.5\cdot 10^{-3})$\\
$q_{2_{pheno_2}}:$ & $5.6\cdot 10^{-3} \leq \text{\texttt{Q}} \leq 1.6\cdot 10^{-2}\ \wedge\ 2.0\cdot 10^{-3} \leq \text{\texttt{VI}} \leq 7.8\cdot 10^{-3}$ \\
$q_{3_{pheno_2}}$ & $1.7\cdot 10^{-3} \leq \text{\texttt{COG}}_{517} \leq 3.3\cdot 10^{-3} \  \wedge\  1.5\cdot 10^{-3} \leq \text{\texttt{COG}}_{201} \leq 2.6\cdot 10^{-3}$\\
 quality: & $J(R_{pheno_2}) = 0.56,\ |supp(R_{pheno_2})| = 5,\ p(R_{pheno_2}) = 3.0\cdot 10^{-12}$ \\
 & \\ 

$R_{pheno_3}:$ & $(q_{1_{pheno_3}},q_{2_{pheno_3}}, q_{3_{pheno_3}})$\\
$q_{1_{pheno_3}}:$ & $1.2\cdot 10^{-5} \leq \text{\texttt{taxID}}_{225194} \leq 1.9\cdot 10^{-3}\ \wedge\  2.7\cdot 10^{-4} \leq \text{\texttt{taxID}}_{384} \leq 4.1\cdot 10^{-3}$ \\
$q_{2_{pheno_3}}:$ & $1.6\cdot 10^{-3} \leq \text{\texttt{IP}} \leq 7.8\cdot 10^{-3}\ \vee (\ 3.5\cdot 10^{-4} \leq \text{\texttt{YL}} \leq 7.2\cdot 10^{-3}\ \wedge\ 4.0\cdot 10^{-4} \leq \text{\texttt{HY}}\leq 6.0\cdot 10^{-3}\ \wedge\ 4\cdot 10^{-4}\leq \text{\texttt{EA}}\leq 3.0\cdot 10^{-3})$ \\
& $\ \vee\ (3.3\cdot 10^{-3} \leq \text{\texttt{PR}}\leq 1.0\cdot 10^{-2})$\\
$q_{3_{pheno_3}}:$ & $\neg (4.2\cdot 10^{-4} \leq \text{\texttt{COG}}_{2885} \leq 4.6\cdot 10^{-3}\ \wedge\ 1.3\cdot 10^{-3} \leq \text{\texttt{COG}}_{317} \leq 2.3\cdot 10^{-3})\ \vee\ (4.5\cdot 10^{-5} \leq \text{\texttt{COG}}_{99}\leq 1.6\cdot 10^{-3}\ $ \\
 & $ \wedge\ 8.1\cdot 10^{-5}\leq \text{\texttt{COG}}_{1534}\leq 9.7\cdot 10^{-4} )$\\
quality: & $J(R_{pheno_3}) = 0.8,\ |supp(R_{pheno_3})| = 68,\ p(R_{pheno_3}) = 1.7\cdot 10^{-3}$\\
& \\

$R_{sw_1}:$ & $(q_{1_{sw_1}},q_{2_{sw_1}}, q_{3_{sw_1}}, q_{4_{sw_1}})$ \\
$q_{1_{sw_1}}:$ & $0.52 \leq \text{\texttt{SiO}}_{2}\leq 1.42\ \wedge\ 0.24 \leq \text{\texttt{Cl}}\leq 0.39\ \wedge\ 0.06\leq \text{\texttt{NH}}_{4}\leq 0.11\ \wedge\ 0.54\leq \text{\texttt{CO}}_{2}\leq 9.56$\\
$q_{2_{sw_1}}:$ & $5 \leq \text{\texttt{49700Sub}} \leq 5\ \wedge\  0 \leq \text{\texttt{19400Sub}}\leq 0$ \\
$q_{3_{sw_1}}$ & $5 \leq \text{\texttt{49700ET}} \leq 5 \  \wedge\  0 \leq \text{\texttt{19400ET}} \leq 0$\\
$q_{4_{sw_1}}$ & $3 \leq \text{\texttt{50390ROS}} \leq 5 \  \wedge\  1 \leq \text{\texttt{49700ROS}} \leq 5\ \wedge\ 0 \leq \text{\texttt{25400ROS}}\leq 0$\\
 quality: & $J(R_{sw_1}) = 0.4,\ |supp(R_{sw_1})| = 16,\ p(R_{sw_1}) = 0.0$ \\
 & \\ 

$R_{sw_2}:$ & $(q_{1_{sw_2}},q_{2_{sw_2}}, q_{3_{sw_2}}, q_{4_{sw_2}})$ \\
$q_{1_{sw_2}}:$ & $0.05 \leq \text{\texttt{NH}}_{4}\leq 8.44\ \wedge\ 0.17\leq \text{\texttt{NO}}_{2}\leq 7.2\ \wedge\ 0.22 \leq \text{\texttt{KMnO}}_{4}\leq 5.88\ \wedge\ 0.22\leq \text{\texttt{Cl}}\leq 6.56$\\
$q_{2_{sw_2}}:$ & $0 \leq \text{\texttt{57500Sub}} \leq 0\ \wedge\ 0 \leq \text{\texttt{50390Sub}}\leq 1\ \wedge\ 1 \leq \text{\texttt{19400Sub}}\leq 5$ \\
$q_{3_{sw_2}}$ & $1 \leq \text{\texttt{19400ET}} \leq 5$\\
$q_{4_{sw_2}}$ & $0 \leq \text{\texttt{57500ROS}}\leq 0\  \wedge\ 0 \leq \text{\texttt{50390ROS}}\leq 1 \ \wedge\ 1 \leq \text{\texttt{19400ROS}} \leq 5$\\
 quality: & $J(R_{sw_2}) = 0.51,\ |supp(R_{sw_2})| = 94,\ p(R_{sw_2}) = 0.0$ \\
 & \\ 
 
 
$R_{sw_3}:$ & $(q_{1_{sw_3}},q_{2_{sw_3}}, q_{3_{sw_3}}, q_{4_{sw_4}})$ \\
$q_{1_{sw_3}}:$ & $0.69 \leq \text{\texttt{bod}}\leq 3.88\ \wedge\ 0.51 \leq \text{\texttt{NH}}_{4}\leq 4.38\ \wedge\ 0.09\leq \text{\texttt{NO}}_{3}\leq 2.15\ \wedge\ 0.23 \leq \text{\texttt{NO}}_2 \leq 7.20$\\
$q_{2_{sw_3}}:$ & $5 \leq \text{\texttt{37880Sub}} \leq 5\ \wedge\ 0 \leq \text{\texttt{17300Sub}}\leq 0$ \\
$q_{3_{sw_3}}$ & $0 \leq \text{\texttt{17300ET}} \leq 0 \  \wedge\  5 \leq \text{\texttt{37880ET}} \leq 5$\\
$q_{4_{sw_3}}$ & $0 \leq \text{\texttt{49700ROS}} \leq 0 \  \wedge\  5 \leq \text{\texttt{37880ROS}} \leq 5$\\
 quality: & $J(R_{sw_3}) = 0.48,\ |supp(R_{sw_3})| = 10,\ p(R_{sw_3}) = 0.0$ \\
 & \\ 
 
 
 $R_{sw_4}:$ & $(q_{1_{sw_4}},q_{2_{sw_4}}, q_{3_{sw_4}}, q_{4_{sw_4}})$ \\
$q_{1_{sw_4}}:$ & $0.48 \leq \text{\texttt{Cl}}\leq 6.56\ \vee\ (0.31 \leq \text{\texttt{Cl}}\leq 0.45\ \wedge\ 0.17\leq \text{\texttt{NO}}_2\leq 2.25\ \wedge\ 0.0\leq \text{\texttt{CO}}_2\leq 0.0 \ \wedge\ 1.91\leq \text{\texttt{O}}_{2}\text{\texttt{sat}}\leq 4.66)$\\
$q_{2_{sw_4}}:$ & $1 \leq \text{\texttt{19400Sub}} \leq 5\ \wedge\ 0\leq \text{\texttt{50390Sub}}\leq 0$ \\
$q_{3_{sw_4}}$ & $3 \leq \text{\texttt{19400ET}} \leq 5$\\
$q_{4_{sw_4}}$ & $1 \leq \text{\texttt{19400ROS}} \leq 5 \  \wedge\  0\leq \text{\texttt{59300ROS}} \leq 1$\\
 quality: & $J(R_{sw_4}) = 0.5,\ |supp(R_{sw_4})| = 94,\ p(R_{sw_4}) = 0.0$ \\
 & \\ 
%


$R_{sw_5}:$ & $(q_{1_{sw_5}},q_{2_{sw_5}}, q_{3_{sw_5}}, q_{4,{sw_5}})$ \\
$q_{1_{sw_5}}:$ & $0.48 \leq \text{\texttt{Cl}}_{2}\leq 6.56\ \vee\ (1.90 \leq \text{\texttt{O}}_{2}\text{\texttt{sat}}\leq 5.92\ \wedge\ 0.03\leq \text{\texttt{NH}}_{4}\leq 0.47\ \wedge 0.46\leq \text{\texttt{NO}}_{3}\leq 6.29\ \wedge\ 0.31 \leq \text{\texttt{bod}}\leq 1.12)\ \vee\ $ \\
& $(0.76\leq \text{\texttt{SiO}}_{2}\leq 2.91\ \wedge\ 0.28\leq \text{\texttt{Cl}}\leq 0.42\ \wedge\ 3.69\leq \text{\texttt{O}}_{2}\text{\texttt{sat}}\leq 4.67\ \wedge\ 1.49\leq \text{\texttt{conduct}}\leq 2.63)$\\
$q_{2_{sw_5}}:$ & $0 \leq \text{\texttt{49700Sub}} \leq 3\ \wedge\ 1 \leq \text{\texttt{19400Sub}}\leq 5$ \\
$q_{3_{sw_5}}$ & $0 \leq \text{\texttt{30400ET}} \leq 0 \  \wedge\  1 \leq \text{\texttt{19400ET}} \leq 5\ \vee\ (0 \leq \text{\texttt{37880ET}}\leq 3\ \wedge\ 0\leq \text{\texttt{49700ET}}\leq 3\ \wedge\ 1\leq \text{\texttt{19400ET}}\leq 5\ \wedge\ 0 \leq \text{\texttt{17300ET}}$ \\
& $ \ \vee\ (1 \leq \text{\texttt{25400ET}}\leq 3)$\\
$q_{4_{sw_5}}$ & $1 \leq \text{\texttt{19400ROS}} \leq 5 \  \wedge\  0 \leq \text{\texttt{59300ROS}} \leq 1 $\\ 
 quality: & $J(R_{sw_5}) = 0.52,\ |supp(R_{sw_5})| = 98,\ p(R_{sw_5}) = 0.0$ \\
 & \\ 
 
 
\end{tabular}
\end{table*}
\end{footnotesize}
\end{center}

Very accurate redescriptions $R_{pheno_1}$ and $R_{pheno_3}$ from Table \ref{tab:redgEx} describe specific ($R_{pheno_1}$) or large ($R_{pheno_3}$) subsets of phenotypes using subsets of features found important by the Random Forest algorithm for predicting these phenotypes. By using the full query language, as in $R_{pheno_3}$, redescription mining can describe very complex relations between important features (as obtained from some predictive model) for some subset of phenotypes. The predictive importance of attributes contained in queries of $R_{pheno_1}$ and $R_{pheno_2}$ for the described phenotypes is low to medium.  $R_{pheno_2}$ contains features with medium to very high predictive importance for the described subset of phenotypic traits. This type of knowledge is not easy to find and it provides useful information about the underlying model (what features does it find predictive for a given subset of phenotypes) and the underlying problem (given an accurate model, further analyses can be made of connections between a given subset of phenotypes and a selected subset of features).

Redescriptions $R_{sw_1}$ - $R_{sw_5}$ from Table \ref{tab:redgEx} demonstrate the use of multi-view redescription mining to explain predictions made by different machine learning models. Redescription $R_{sw_1}$ describes $16$ water samples with the Jaccard index $0.4$. This means that there are additional $24$ water samples described with a subset of queries but not all of them. This redescription reveals that all models predict the occurrence of species $49700$ (Gammarus Fossarum) on the locations contained in its support set, two out of three queries predict no-occurrence of species $19400$ (Nitzschia Palea) on these locations and the ROS model predicts moderate to high occurrence of species $50390$ (Baetis Rhodani) and no occurrence of species $25400$ (Cladophora). Domain expert can immediately see what properties hold from the descriptive attributes (physical and chemical measurements). What is interesting, and not easily derivable by only looking at the predictions made by these models is that as species $49700$ seems abundant (or at least present), species $19400$ seems to be predicted mostly absent. Indeed, by checking the selected redescription with the ground truth (real target labels of entities from the support set of this redescription), $13$ out of $16$ locations contain species $49700$ in abundance ($5$), $3$ locations contain medium occurrence of species $49700$ ($3$). Thus, all models rightfully predicted occurrence of this species in all redescribed locations, and mostly even the abundance level. $15$ out of $16$ locations do not contain occurrence of species $19400$ which is in a large accordance with predictions of the Random Subsets and Extra multi-target PCTs. Species $25400$ is also mostly absent ($12$ out of $15$ locations) whereas species $50390$ has medium to abundant occurrence ($3$ or $5$) in $12$ locations and rare occurrence in $2$ locations. By using their domain knowledge, a domain expert may choose to trust only a subset of models or use some compromise as prediction (which requires examining predictions of base classifiers). When this is done, redescription predictive quality measures can be computed in the same way as for any other classification algorithm.

The redescription $R_{sw_2}$ contains properties of one large cluster of locations. As predicted by all models, these $94$ locations should contain at least small presence of species $19400$ (Nitzschia Palea). Ground truth target labels show that $82$ out of $94$ locations indeed have at least small presence of this species and $58$ medium to high presence. Checking real accuracy of this redescription requires checking predictions of underlying models (which disagree on substantial subset of locations), mostly, there are many substitutions of neighbouring classes $0\leftrightarrow 1,\ 1\leftrightarrow 3$, $3\leftrightarrow 5$ that occur in one or more underlying models. A smart way of joining these models into an ensemble, as using the obtained rules, may potentially yield higher accuracy than that obtained individually by the base models. $74$ out of $94$ locations have no occurrence of species $50390$ (Baetis Rhodani) and $84$ out of $94$ locations have no occurrence of species $57500$ (Rhyacophila).


Identifying subsets of entities on which multiple models agree or have very good performance is not the only benefit of this approach, since it can also be used to detect and analyse problematic subsets (these on which used models make mistakes or disagree upon). The redescription $R_{sw_3}$ redescribes $10$ locations for which all models made significant classification errors. Although all models predicted abundant occurrence of species $37880$ (Tubifex), ground truth target labels show that only $5$ locations have medium occurrence of this species. The species $17300$ (Melosira Varians) which is predicted to have no occurrence in  locations from support set of redescription $R_{sw_3}$ has small occurrence on $5$ locations and the species $49700$ (Gammarus Fossarum) has small occurrence on $2$ locations.

When using disjunctions, redescriptions can contain complex descriptions of different locations with the different measurements (as in $R_{sw_4}$) or of occurrence and partial co-occurrence of different species (as in $R_{sw_5}$). Although very complex, such redescriptions may be used as a general filter to identify entities with some complex property.

Technique for understanding machine learning models presented in this section and its capabilities significantly differ from the capabilities of the well known methods for explaining predictions such as  SHAP \cite{SHAP} or LIME \cite{LIME}. These methods aim to explain why some supervised machine learning model made the obtained predictions. It does that either by learning an interpretable model locally around the prediction (LIME) or by assigning a importance value to each feature for a particular prediction (SHAP). 


\section{Conclusion and future work}
\label{conclusions}

The main goal of this work is to present a general, memory efficient framework for multi-view redescription mining. The proposed framework starts by creating two-view redescriptions with the GCLUS-RM algorithm. The main idea behind our efficient multi-view redescription mining algorithm is to use redescriptions as targets in the redescription completion phase, which extends two-view redescriptions to multiple views. This significantly reduces the computational complexity of the approach. 

Comparison results with the naive extension of $2$-view redescription mining algorithms to multi-view setting demonstrate that unlike the naive extension, the proposed methodology is generally applicable, it mostly outperforms the naive extensions with respect to redescription accuracy, significance, overall redescription set score, memory consumption and execution time and it heavily outperforms the naive extensions with respect to all these measures on datasets that allow creating large number of redescriptions. Detailed evaluation of the proposed framework revealed: 
\par \vspace{1mm}\noindent
1) Using a larger amount of memory tends to increase the overall redescription quality. Increasing the number of iterations inside the GCLUS-RM additionally increases the difference in quality between redescription sets produced using low and high amounts of memory. 
\par \vspace{1mm}\noindent
2) Using a supplementing model significantly increases the redescription accuracy, diversity, and overall number of produced redescriptions satisfying some predefined quality constraints. This leads to redescription sets with superior properties as compared to those obtained using one rule-generating model obtained using the PCT algorithm. 
\par \vspace{1mm}\noindent
3) It is feasible to use the Extra multi-target PCTs algorithm to create the main rule-generating model and that, with $4$ trees or more, such an approach outperforms the use of a rule-generating model obtained using one PCT. This is important, because using the Extra multi-target PCTs algorithm to produce the main rule-generating model reduces the overall computational complexity of the approach. 
\par \vspace{1mm}\noindent
4) Using the random subspace view projections may be a feasible approach to reduce the computational cost of the framework or to perform preliminary prototyping. Although the produced sets have a slightly lower score (and are more susceptible to random fluctuation), using projection allows significant execution time savings (up to $3$ times when $4$ views are available). 
\par \vspace{1mm}\noindent
5) There are large benefits in incorporating different information from predictive models (such as feature rankings or predictions) into multi-view redescription mining. 
\par \vspace{1mm}\noindent
6) Multi-view redescription mining can be used to increase the overall understanding of the studied problem domain and the predictive models used to obtain predictions or feature rankings.

Currently, the approach computes the Cartesian product of two rule-sets to obtain redescriptions. Interesting direction for future work includes efficiently reducing the number of tests needed to compute this set. Other direction includes discovering ways to choose a subset of initial views, in a guided manner, that will produce the maximal number of redescriptions, when random view subspace projection is used. Contributions along these two directions will surely enhance the efficiency of the approach. 

\section*{Acknowledgment}
We thank prof. Sašo Džeroski for reading and commenting one of the earlier versions of the manuscript. This work was supported in part by the: “Research Cooperability“ Program
of the Croatian Science Foundation funded by the European Union from the
European Social Fund under the Operational Programme Efficient Human
Resources $2014$-$2020$, grant $8525$ "Augmented intelligence workflows for
prediction, discovery and understanding in Genomics and
Pharmacogenomics", and European Regional Development Fund under the
grant KK$.01.1.1.01.0009$ (DATACROSS).

\begin{IEEEbiography}[{\includegraphics[width=1in,height=1.25in,clip,keepaspectratio]{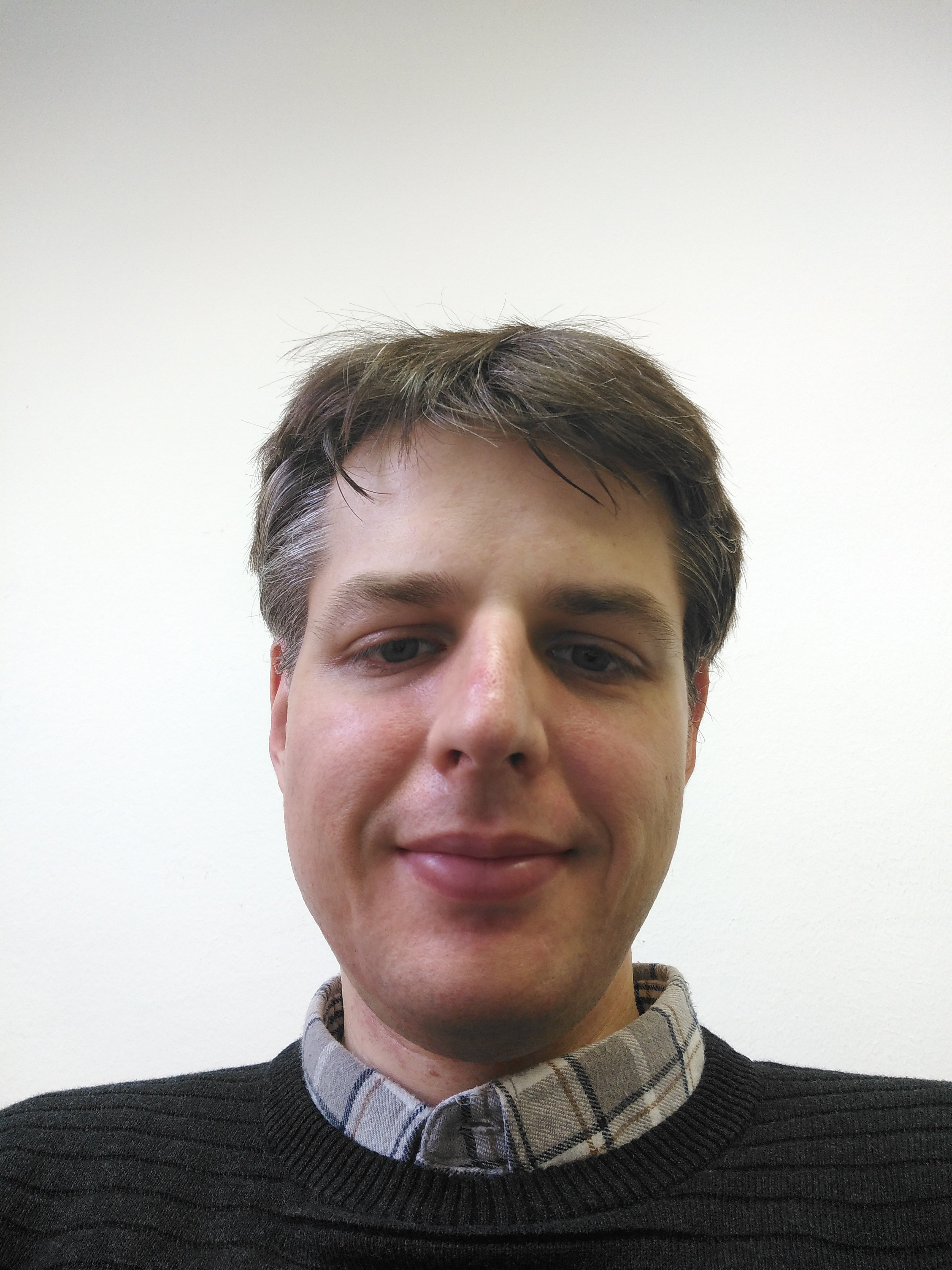}}]{Matej Mihel\v{c}i\'{c}} received a B.S. in Mathematics in $2009.$ from the Department of Mathematics, Faculty of Science, University of Zagreb, Croatia. He obtained the M.S. in computer science and mathematics from the  same institution in $2011.$ and the PhD diploma in computer science (thesis topic: \emph{Construction and Exploration of Redescription Sets}) from the International Postgraduate School Jožef Stefan, Ljubljana, Slovenia in $2018$. 

During his PhD study ($2013. - 2018.$), he was working as a Research Assistant at the Ruđer Bošković Institute in Zagreb, Croatia and was a visiting PhD student at the Research group for Genome Data Science, Institute for Research in Biomedicine, Barcelona, Spain ($2018.$).  He  worked as a Postdoctoral Researcher at the School of Computing, Faculty of Science and Forestry, University of Kuopio, Kuopio, Finland ($2019. - 2020.$). He is currently working as a Postdoctoral Researcher, Teaching Assistant and Lecturer at the Department of Mathematics, Faculty of Science, University of Zagreb, Zagreb, Croatia. He has co-authored one Book Chapter and $14$ scientific manuscripts published in different computer science or multidisciplinary scientific journals, and proceedings of international conferences. His main research field is redescription mining, however his interests include development of novel techniques, algorithms and tools for data analyses, knowledge discovery - primarily in biology and medicine, decision support, interpretable data mining and machine learning.  He served as a reviewer for one scientific journal and two international conferences. 
\end{IEEEbiography}

\begin{IEEEbiography}[{\includegraphics[width=1in,height=1.25in,clip,keepaspectratio]{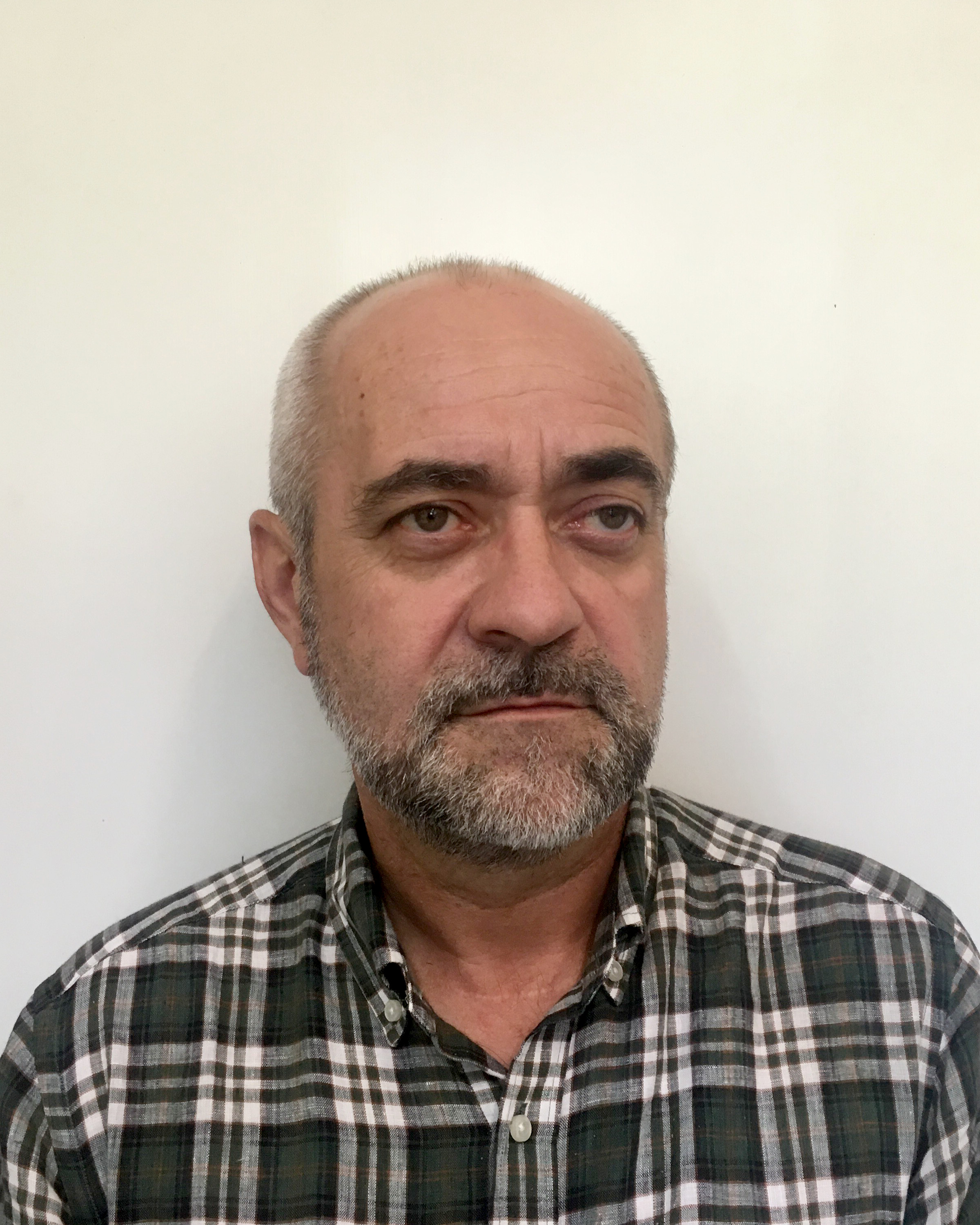}}]{Tomislav \v{S}muc} PhD is a Head of Laboratory for Machine Learning and Knowledge
Representation at Ruđer Bošković Institute, Zagreb. His research interest is in the
area of artificial intelligence, in development and application of machine learning
and data mining techniques for knowledge discovery in different domains of
science and technology. He has been participating in, or leading, a number of
research projects financed by Croatian, European and other international funding
agencies. TS was mentor of a dozen of master's and PhD students at the University of
Zagreb and was involved in organization of a number of international conferences
(ECML-PKDD, Discovery Science), workshops and summer schools. Tomislav Šmuc has published over
$100$ papers in journals and proceedings of international conferences and serves as a reviewer for a number of scientific journals in the fields of computer science, computational biology and interdisciplinary science.
\end{IEEEbiography}

\EOD

\end{document}